\documentclass[10pt,letterpaper]{article}
\usepackage[top=0.85in,left=2.75in,footskip=0.75in]{geometry}
\usepackage{amsmath,amssymb}
\usepackage{changepage}
\usepackage[utf8x]{inputenc}
\usepackage{textcomp,marvosym}
\usepackage{cite}
\usepackage{nameref,hyperref}
\usepackage[right]{lineno}
\usepackage{microtype}
\DisableLigatures[f]{encoding = *, family = * }
\usepackage[table]{xcolor}
\usepackage{array}
\usepackage{mathtools}
\usepackage{hyperref}
\hypersetup{colorlinks=true,linkcolor=blue}

\DeclarePairedDelimiter\floor{\lfloor}{\rfloor}

\usepackage{pdfpages}

\newcolumntype{+}{!{\vrule width 2pt}}
\newlength\savedwidth

\raggedright
\usepackage[aboveskip=1pt,labelfont=bf,labelsep=period,justification=raggedright,singlelinecheck=off]{caption}

\makeatletter
\renewcommand{\@biblabel}[1]{\quad#1.}
\makeatother
\usepackage{lastpage,fancyhdr,graphicx}
\usepackage{epstopdf}
\usepackage{tikz}
\fancyheadoffset[L]{2.25in}
\fancyfootoffset[L]{2.25in}
\begin{document}
	\vspace*{0.2in}
	
	\begin{flushleft}
		{\Large
			\textbf\newline{Compatible deep neural network framework with financial time series data, including data preprocessor, neural network model and trading strategy} 
		}
		\newline
		\\
		Mohammadmahdi Ghahramani\textsuperscript{1\Yinyang},
		Hamid Esmaeili Najafabadi\textsuperscript{2\ddag}
		\\
		\bigskip
		\textbf{1} Mathematics Department/University of Padova, Padua, Italy
		\\
		\textbf{2} Electrical and Computer Engineering Department/University of Calgary, Calgary, Canada
		\\
		\bigskip
		
		%
		%
		\Yinyang muhammadmahdi0007@gmail.com
		
		\ddag hamid.esmaeilinajafa@ucalgary.ca
		

	\end{flushleft}
	
	\section*{Abstract}
	Experience has shown that trading in stock and cryptocurrency markets has the
	potential to be highly profitable. In this light, considerable effort has been recently
	devoted to investigate how to apply machine learning and deep learning to interpret
	and predict market behavior. This research introduces a new deep neural network
	architecture and a novel idea of how to prepare financial data before feeding them
	to the model. In the data preparation part, the first step is to generate many features
	using technical indicators and then apply the XGBoost model for feature engineering.
	Splitting data into three categories and using separate autoencoders, we extract
	high-level mixed features at the second step. This data preprocessing is introduced to
	predict price movements. Regarding modeling, different convolutional layers, an long short-term memory unit, and several fully-connected layers have been designed to perform binary
	classification. This research also introduces a trading strategy to exploit the trained
	model outputs. Three different datasets are used to evaluate this method, where
	results indicate that this framework can provide us with profitable and robust
	predictions.

	\section*{Introduction}
	Acting on cryptocurrency and stock markets has recently gained a great deal of popularity. People's curiosity about financial keywords such as Bitcoin (BTC), Ethereum (ETH), Tether (USDT), cryptocurrency, technical analysis, etc., has increased significantly. There has been a 200 percent increase in this interest for some Asian countries during the last year as per google trends data. Apart from people's willingness, researchers have also been seeking to develop efficient methods to deal with the complexity of financial markets. Focusing more on modeling, a significant number of studies were conducted on how to predict the price movement direction using machine learning algorithms and deep neural networks (DNNs). Thanks to advances in neural networks, the dream of forecasting financial time series is now much closer to reality.

	Generally speaking, when it comes to financial data, there are two main problems we should address. The first one is to introduce robust features through which the market can be correctly represented~\cite{chong2017deep} and the second one is market complexity and volatility~\cite{tsantekidis2017using}. Because of market elements' nonlinear relationships~\cite{maknickiene2011investigation, chen2020financial, bao2017deep}, it is not surprising deep neural networks are used a lot to predict market fluctuations. They vary from using fully-connected neural networks~\cite{vahala2016prediction} to taking advantage of convolutional neural networks (CNNs)~\cite{wang2018financial, hoseinzade2019u, hoseinzade2019cnnpred, hoseinzade2018cnnpred}, recurrent neural networks (RNNs)~\cite{tsantekidis2017using, samarawickrama2017recurrent} and long short-term memory (LSTM)~\cite{zanc2019forecasting, fischer2018deep, zhelev2019using}. However, the critical point would be that instead of spending too much time designing complex neural network architectures, we should equally care about modeling and extracting valuable features.
	
	This study is divided into two parts: data preparation and modeling. This research tries to provide intuition for each step to make it closer to what is happening in the stock or cryptocurrency markets. 
	
	Feature generation (FG), data denoising (DD), feature engineering (FE), data splitting (DS), create representative dataset (CD), and using auto-encoders (AE) are considered as six crucial steps for the first part. Modeling part also involves convolutional layers, LSTM units, and fully-connected layers for binary classification. Since the concern of robustness in financial tasks has always been an issue, the performance of the proposed method is evaluated on three different datasets. They include four-hour data of BTC and ETH coins from the cryptocurrency market and daily data for the S\&P500 index.
	
	For a trader, it is imperative to use technical analysis according to the current market state correctly. Traders cannot afford to use only one specific indicator from which they might have continuously gained profit, e.g., exponentially moving average (EMA). Although it may have brought them significant profit in one of the market states, it immediately loses its performance as the market state changes.  They also need to utilize other tools, such as price channels or other indicators like relative strength index (RSI). It is indeed the unique quality distinguishing amateur traders from professionals. Taking advantage of this intuition, it is quickly unfolded that feature engineering should be dynamic. Gathering a set of features to represent all or most corresponding states not only for different markets but also for different seasons of the single market is critical.  In this light, we first generate numerous potential features from open, high, low, close prices and volume data (OHLCV) to design a dynamic feature extractor. Then, we apply a wavelet transformer to denoise the generated features~\cite{bao2017deep, struzik2001wavelet}. Finally, feature engineering is applied to generated features since the importance levels of these features are still unknown~\cite{nobre2019combining}.
	
	Neural network models are notably susceptible to features quality. If they are fed with low-quality features, their performance is less likely to be as desired. Here, the idea of using auto-encoders is introduced to ensure that only high-quality features are fed to our primary model. They have been used in other studies to extract both high-quality features~\cite{bao2017deep} and data denoising~\cite{turkmen2015application}. However, the approach used in this paper is totally different from these studies since we designed customized CNN-based auto-encoders~\cite{masci2011stacked} and used two auto-encoders instead of one. An auto-encoder aims to learn a compressed, distributed representation for a dataset or reduce data dimensionality. It can give us the edge to make the proposed model work better. Auto-encoder's performance depends on feature correlation such that a higher degree of correlation between the input features boosts the auto-encoder's performance. When it comes to financial markets, features are not necessarily correlated enough. Hence, feeding them all directly into an auto-encoder would not result in extracting high-quality features. To address this challenge, we consider the data splitting step. Before feeding these splitted data to separate auto-encoders, it is worth deliberating on  how traders are trading and whether they only look at one single candle or they tend to review the price path when deciding to buy or sell. This intuition tells us a single candle should not be considered a training instance but instead their trend. In this light, a new dataset is created in step CD, where candle sets are considered training instances. The data is now ready to be fed to auto-encoders. The rest of this paper is about designing a neural network-based architecture to binary-classify the prepared data coming from the preparation part. Even though some studies attempted to predict the value of price directly using regression methods and then define a trading strategy to exploit results~\cite{bao2017deep}, it seems it would be more efficient and closer to reality to predict the price movement direction instead of the price value.
	
	This paper considers a trading strategy part that determines how this model should be applied in our financial activities to exploit our model outputs efficiently. This part has been rarely considered in other studies. It is precisely why they mainly can not be used by organizations and traders to earn profit from financial markets. It first explains several unrealistic assumptions which may make us wrongly evaluate our model and then suggests intuitive solutions to address them. Secondly, it talks about indexes to correctly assess the model because coming up with a highly accurate model does not necessarily result in earning profit from the trading system~\cite{zanc2019forecasting}. In contrast, none of all low-accurate models are useless. 
	
	\section*{Related works}
	Having access to the high performance GPUs and hardware along with advances in deep learning and machine learning, the opportunity to utilize all deep learning methods and techniques has been provided for researchers. Consequently, standard deep learning methods that are typically used in specific areas have been tested on financial time series data. J Vahala et al.~\cite{vahala2016prediction} review the performance of fully-connected layers to predict financial markets. Focusing on feature extraction, Xiao Ding et al.~\cite{ding2015deep} use a deep convolutional network to improve prediction accuracy. Using the same concept, Jia Wang et al.~\cite{wang2018financial} propose one-dimensional convolutional neural networks with the aim of market movement forecasting. With respect to the nature of financial data, RNNs and LSTM units are also widely used. Using these networks, Razvan Zanc et al.~\cite{zanc2019forecasting} conclude that accuracy is not a good indicator to evaluate the model profitability, and high values of accuracy may not contribute positively to a profitable trading system. In another paper, Avraam Tsantekidis et al.~\cite{tsantekidis2017using} attempt to exploit RNNs on time-series data on limit order books. Justin Sirignano et al.~\cite{sirignano2019universal}, mark that there is path-dependence in price dynamics. Wei Bao et al.~\cite{bao2017deep} utilize stacked auto-encoders in the combination of LSTM units to predict one-step-ahead close price . In another work by Ali Caner Türkmen et al.~\cite{turkmen2015application}, an array of features are generated, and the performance of several machine learning algorithms are tested.
	
Even though various innovative ideas have been implemented in this field, financial data preprocessing and taking advantage of traders' intuitions are not adequately considered. This paper covers this topic by considering a dynamic feature engineering method. In related research, the intuition behind selecting a method for modeling was explained. However, none of them has tried to select the model's parameters according to the common rules in the market and among traders. As the second edge, not only does this research make an effort to construct an intuitive network, it also suggests setting the network's parameters intuitively. While the importance of existing a trading strategy is not undeniable by traders, defining a trading strategy is another topic that other researchers did not consider. This work also proposes a strategy to turn the model's output into profit.
	
	\section*{Materials and methods}
	\section{Data preparation}
	Providing an informative set of data has always played a vital role in machine learning tasks. The focus of this part is that gathering and engineering data for the market is as crucial as designing complex neural networks. One of the trader's primary tools to analyze the price movement is OHLCV data. Interaction between these features helps market actors to predict upward or downward trends in advance. Based on their importance, we start with a raw dataset including OHLCV data.  
	
	\subsection{Feature generation}
	Financial markets are renowned for their complexity~\cite{arthur1995complexity, brunnermeier2009complexity}. This complexity is coming from a volatile supply and demand pressure which determines the price. Consequently, traders encounter market states where price behavior is unpredictable since different technical tools might provide them with opposite signals. This condition gets even more challenging when traders need to struggle with setting hyperparameters for their technical tools. Consider that it is desired to earn five percent worth of profit by buying Bitcoin, but there is an uncertainty about whether this is the right time. One possible approach is to take a look at the BTC/USDT chart and analyze the price. Then, it must be figured out which set of indicators are better to use. Having done that, optimum parameters for the selected indicators should be set. For example, an EMA indicator with the window of 21 may provide a better prediction rather than the window of 35. To take this fact into account, this part generates too many indicators with different parameters whose summary is brought in Table~\ref{table1}. Therefore, the new dataset may include several EMA columns, each with its own window. Consequently, the variety in both indicators and parameters is provided.
	
		\begin{table}[!htb]
		
		\begin{adjustwidth}{-1.0in}{1in}
			\caption{
				{\bf List of some common technical indicators to generate features}}
			\includegraphics[width=\textwidth]{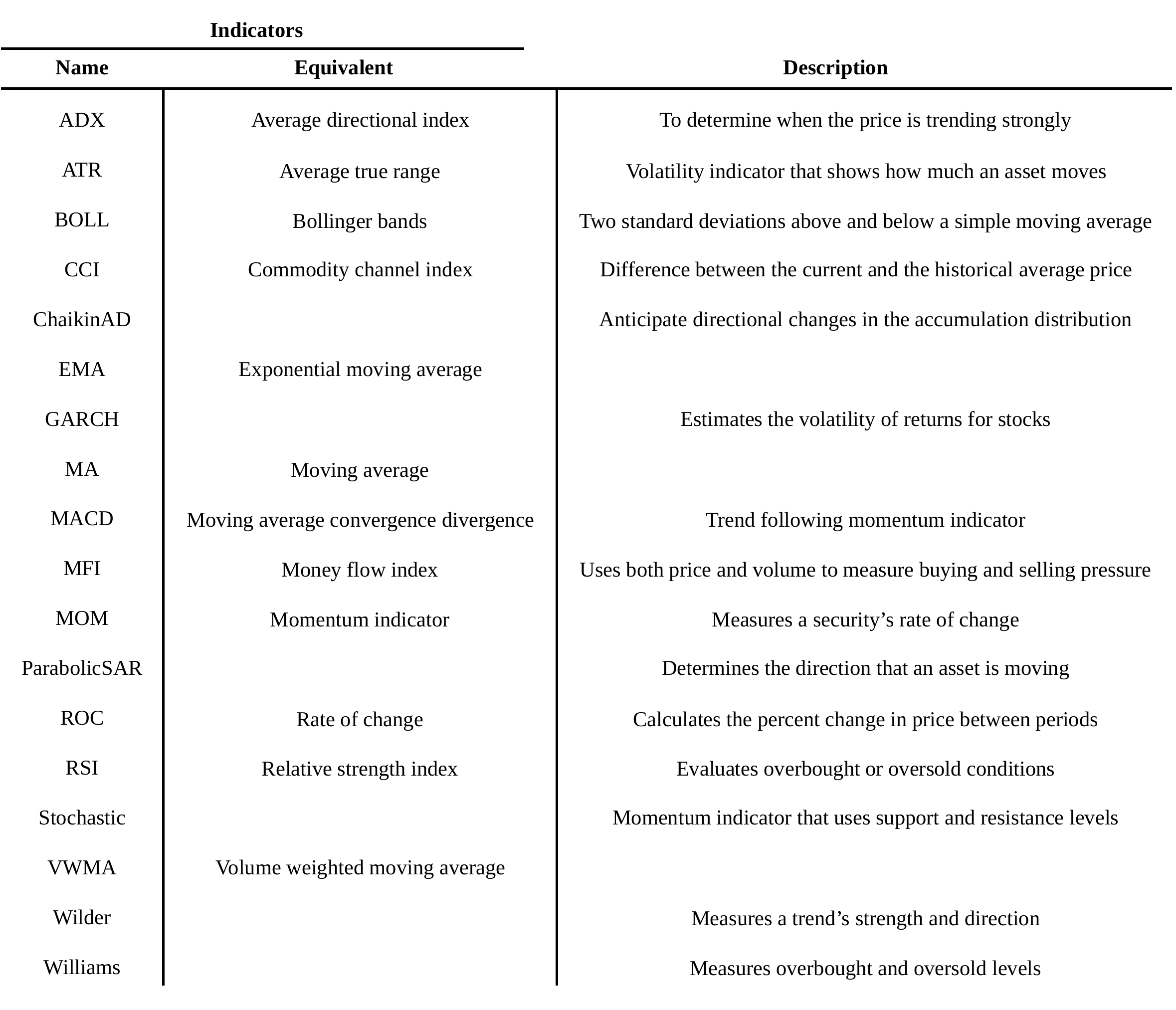}
			
			\begin{flushleft} \small Table notes that various indicators and oscillators are considered to cover common approaches indeed when analyzing financial markets. 
			\end{flushleft}
			\label{table1}
		\end{adjustwidth}
	\end{table}
	
	\subsection{Data denoising}
	People use exchanges to trade in financial markets. Anyone can trade on them, whether they are professionals or novices. As a result, financial markets get noisier than what they already are. All market activities affect demand and supply pressure and, as a result, price. It is important to consider only transactions that contains information. This research suggests taking the data denoising step. However, handling all kinds of noises in financial markets is impossible.

This paper recommends using the wavelet transforms to convert the data to a smoother version of that~\cite{pan1999two}. Wavelet transformers are widely-used to signal denoising, especially in electrical engineering. Since we can suppose our time-series data like a signal, this transformer can be employed with the aim of denoising~\cite{bao2017deep}. Wavelet analysis was first developed in the early 1980s in seismic signal analysis in the form of an integral transform with a localized kernel function with continuous parameters of dilation and translation. They have been applied to various kinds of one-dimensional or multidimensional signals, for example, to identify an event or to denoise the signal. Apart from data denoising, applying wavelet transforms protects the model against over-fitting.

	\subsection{Feature engineering}
	One of the most essential steps for every machine learning task is feature engineering. Feature engineering is the process of using domain knowledge to extract features from raw data. This part cares about transforming raw data into components that better represent the underlying problem to the predictive models, resulting in improved model accuracy on unseen data.

Till now, we generated features and applied wavelet transforms with the aim of denoising. These two basic steps are not adequate to prepare the data. Adding several indicators to a price chart is simply the process of feature generation which even amature traders can do. They may also ignore some upper or lower shadows of candles and imagine the smoother version of the chart. Since the mere use of these steps do not equip them with robust and dependable forecasting, no wonder if the FG and DD steps alone do not give us a significant edge. This part introduces a reliable method to extract valuable features. The paper divides the FE step into two sections. The first one is explained here, and the second one is covered on the auto-encoder part.

\begin{figure}[!htb]
		\begin{adjustwidth}{-1.0in}{1.0in}
			
			\includegraphics[width=\textwidth]{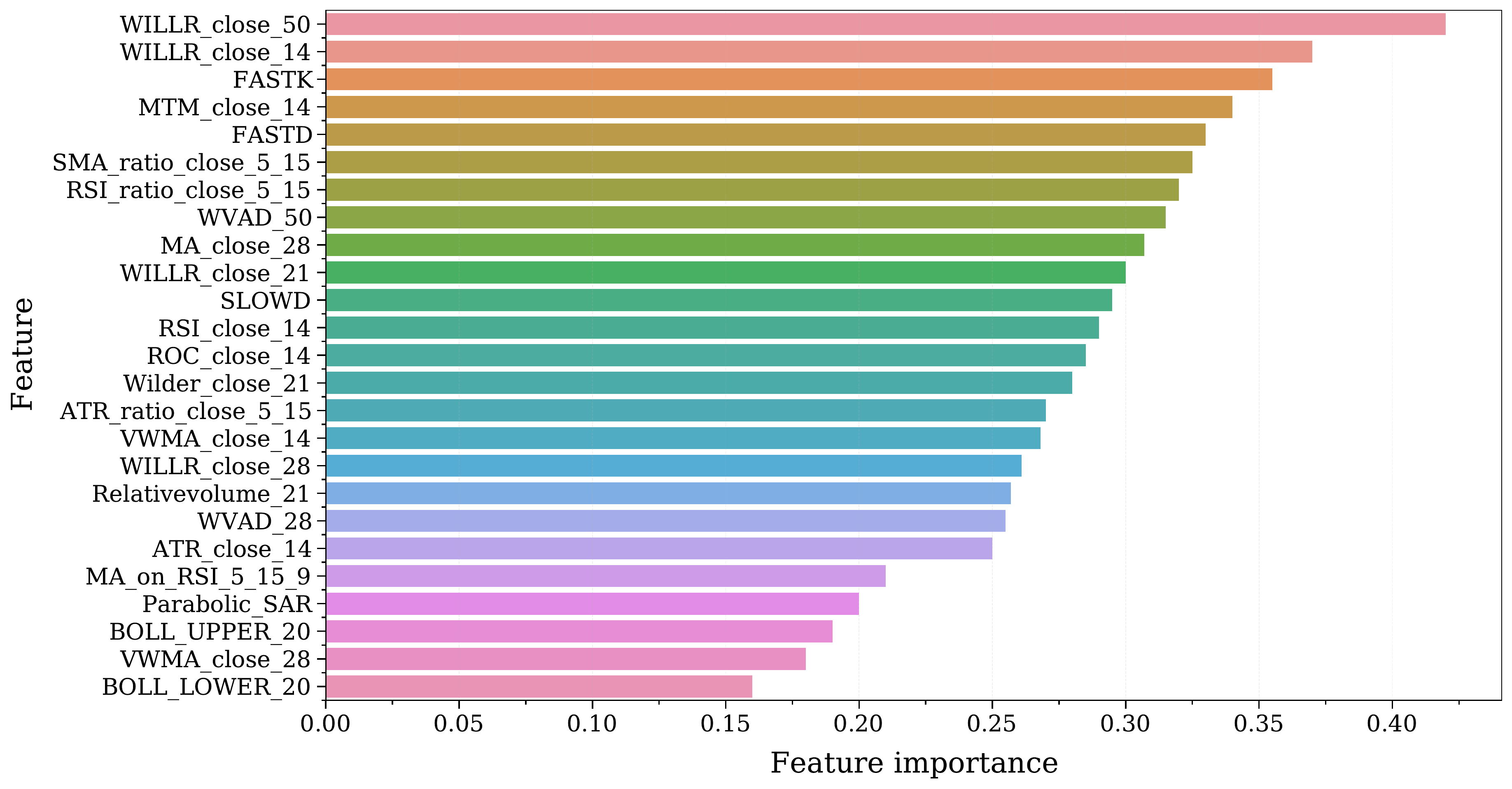}
			
		\caption{\small{
			{\bf Feature importance produced by XGBoost classifier.} The figure illustrates the top 25 features, ranked according to their importance produced by XGBoost for daily data of the S\&P500 market.}}
		
			\label{fig1}
		\end{adjustwidth}
	\end{figure}

The market contains different complex states and we must use a dynamic feature engineering method to represent them. To design that, this study takes advantage of another intuition. Financial market participants tend to consult with experts and use their insights before investing in financial markets. In our case, there is a machine learning method whose primary duty is to determine feature importance. This paper selects XGBoost classifier~\cite{shi2019feature} to play an expert role since tree-based machine learning algorithms are not sensitive about the quality of every single feature and can ignore those features with a lower information~\cite{chen2016xgboost}. This feature turns the XGBoost classifier into an excellent candidate to calculate feature importance. Each time it is fed with different data, feature importance changes. Fig~\ref{fig1} is a snapshot that illustrates how XGBoost's output looks like. Having access to the importance of features, we can simply select the top 25 features, for example, and reduce the number of columns to those whose importance have been endorsed by the XGBoost model.
	
	\subsection{Data splitting}
	Before proceeding with auto-encoders and finishing the data preparation part, this research ensures that the auto-encoders are fed as expected. Auto-encoders expect to receive an adequately correlated set of features, but current XGBoost-passed features are not necessarily correlated enough with each other.
	
	To address this challenge, this study suggests splitting the dataset into correlated bunches of features. It categorizes columns into three main groups named OHLCV, price-like, and non-price-like. The OHLCV group, as its name suggests, includes open, high, low, close, volume columns. The rest of the features are also split into two groups. By price-like group, we refer to a group whose features' unit is the same as price unit. In a nutshell, they can be plotted on a price chart. This group may include different kinds of moving average indicators or any other features which are reported in dollars. By contrast, the last group, a non-price-like group, has features whose units are not the same as price units. This group may include oscillators like RSI and CCI, which are dimensionless. Since we would not like to lose any information from our primary group, the OHLCV data is not passed through an auto-encoder. However, two separate auto-encoders are designed for both price-like and non-price-like groups to extract high-level features.
	
	\subsection{Create Representative Dataset}
	This section provides an intuitive method to form our training dataset. Price path is much more important than the current candle information. The fact that technical analysis such as candlestick patterns is all about tracking the price path is substantial evidence on why this study considers this step.

\begin{figure}[!htb]

		\begin{adjustwidth}{-1.0in}{1.0in}

			\includegraphics[width=\textwidth]{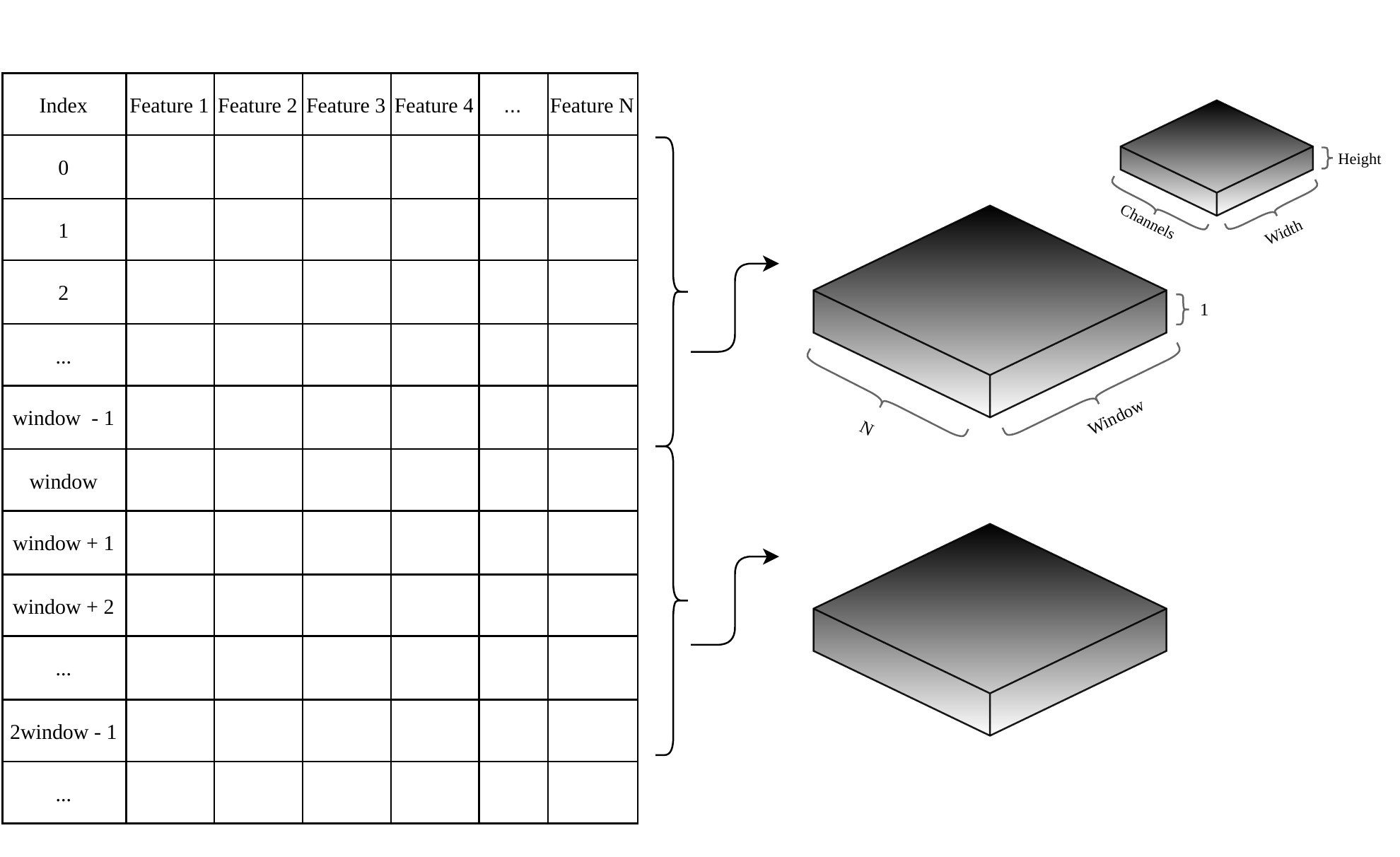}
		
			\caption{\small{
				{\bf Define training instances.}}}
			
			\label{fig2}
		\end{adjustwidth}
	\end{figure}

This research suggests defining each training instance as a set of previous consecutive candles. For example, if we set training samples to consist of 24 candles, the dimension for every sample would be \texttt{number of features$\times$1$\times$24}, representing the number of channels, height, and width, respectively. Fig~\ref{fig2} illustrates how training instances are being created from the dataset. These samples will soon be used to feed the auto-encoders.

	\subsection{Auto-encoder}
	This part talks about the architecture of auto-encoders whose job is to extract high-level features. Their usage is not limited to feature extraction and can be employed to, for example, predict bankruptcy~\cite{soui2020bankruptcy}. As mentioned earlier, this section is the second step associated with feature engineering. The study designs two auto-encoders, one for price-like features and another one for non-price-like features.

\begin{figure}[!htb]

		\begin{adjustwidth}{-1.0in}{1.0in}
			
			\includegraphics[width=\textwidth]{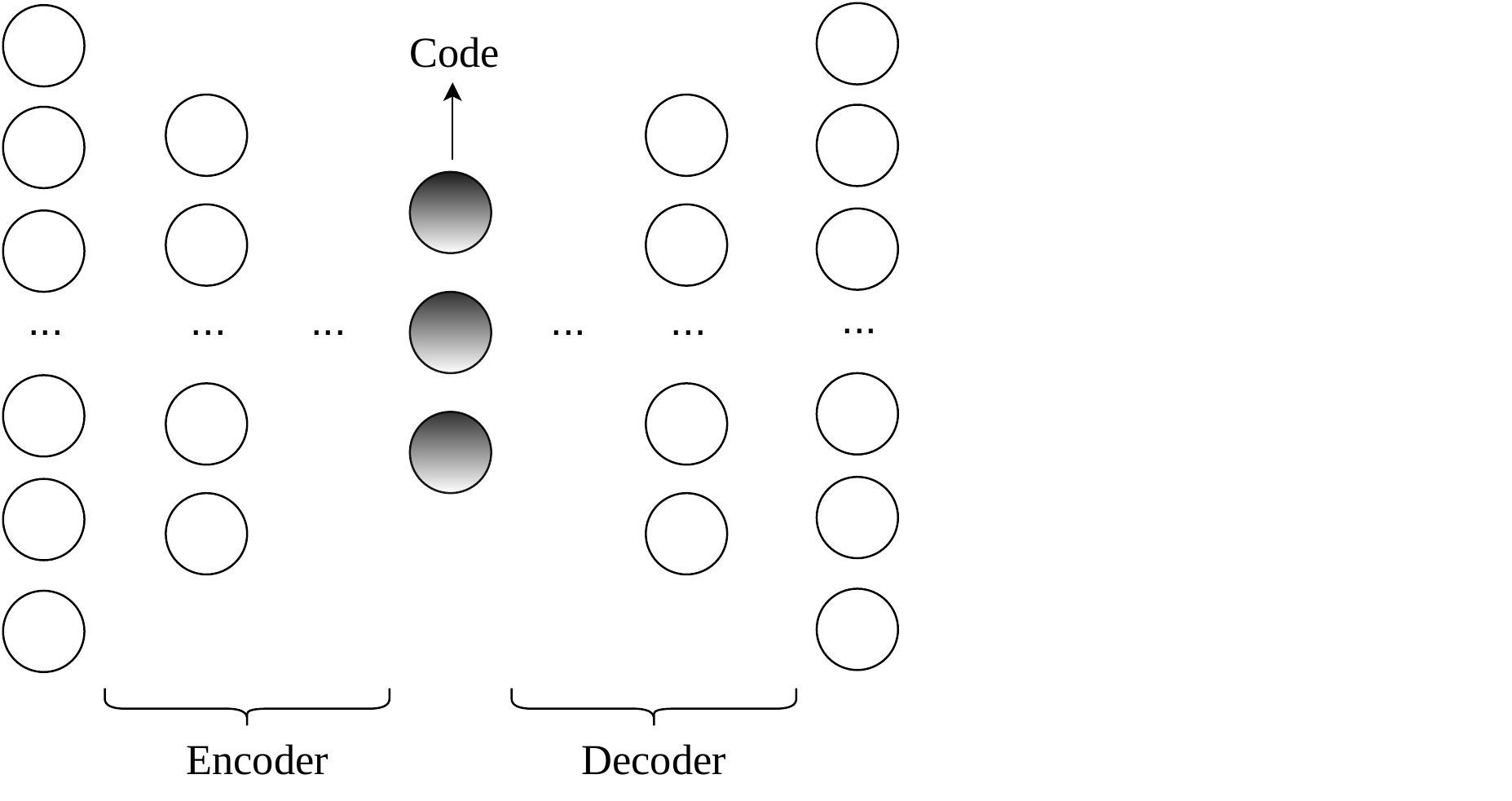}
		
		\caption{\small{
				{\bf Typical auto-encoder.} The figure illustrates three parts of an auto-encoder. The input and output data are the same as each other. The encoder layer converts the input data into the code. Decoder layer also reconstructs the input data using the code. Having auto-encoder trained, the code, which represents the input data, is used instead of the input data for further processes.}}
			
			\label{fig3}
		\end{adjustwidth}
	\end{figure}

As their name suggests, auto-encoders are a specific type of neural network where the input is the same as the output. This kind of learning is categorized as an unsupervised technique since they do not need explicit training labels. More precisely, they are self-supervised because they generate their own labels from the training data. Auto-encoders attempt to reconstruct the input data such that there is a minimum difference between the original input data and the reconstructed output. An auto-encoder consists of three components named encoder, latent-space representation, and decoder. The encoder compresses the input and produces the latent-space representation, also called \texttt{code}. Decoder then reconstructs the input only using this code. The applications of auto-encoders mainly include dimensional reduction~\cite{wang2016auto}, latent space extraction as the high-level compressed features and anomaly detection~\cite{sakurada2014anomaly}. It is essential to feed auto-encoders with correlated data. This paper cares about this correlation and takes some measures to correctly consider this fact. Fig~\ref{fig3} shows the most typical version of auto-encoders, including encoder, code, and decoder.
	
	Even though most auto-encoders are designed as in Fig~\ref{fig3}, this study constructs the auto-encoders with convolutional layers since the shape of input data requires such layers both for the encoder and for the decoder. Instead of hidden neural layers, we use one-dimensional convolutional layers (Conv1d layers) to encode the input data and decode the code. This paper uses the same auto-encoder architecture for price-like and non-price-like features applied separately on their corresponding features.
	\subsubsection{Encoder}
	
\begin{figure}[!htb]

		\begin{adjustwidth}{-1.0in}{1.0in}
			\includegraphics[width=\textwidth]{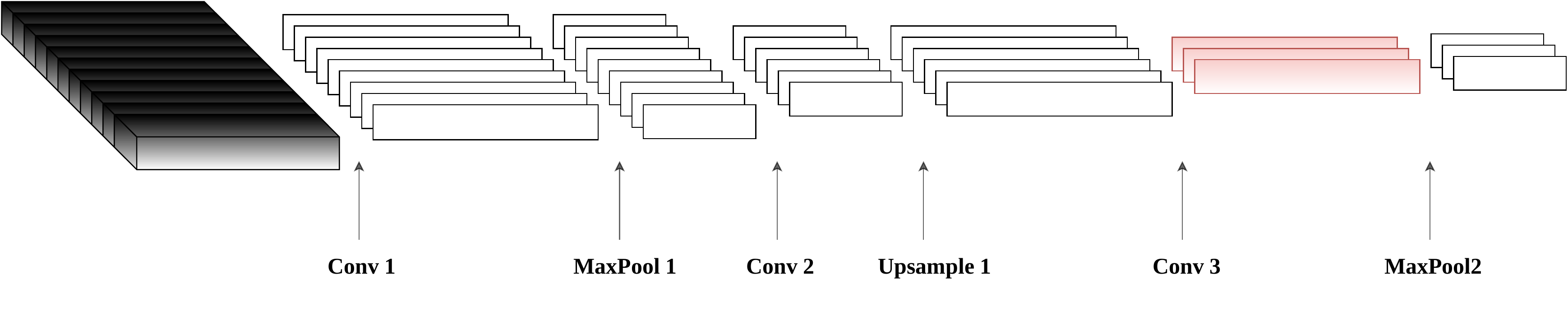}
		
		\caption{\small{
				{\bf Encoder architecture.} This figure considers the code inside the encoder layer just for illustration purposes.}}

			\label{fig4}
		\end{adjustwidth}
	\end{figure}

	Encoder consists of Conv1d layers along with max pool and up sampling layers. The encoded data, red-colored in Fig~\ref{fig4}, is the same in height and width as the input data and the encoder only reduces the channel size. Max pool and up sampling layers make auto-encoder learn better. Consequently, this paper incorporates up sampling layers in addition to max pool layers. More importantly, having the same dimension in width and height requires us to use both max pool and up sampling layers to negate each other.
	
\begin{figure}[!htb]

		\begin{adjustwidth}{-1.0in}{1.0in}

			\includegraphics[width=\textwidth]{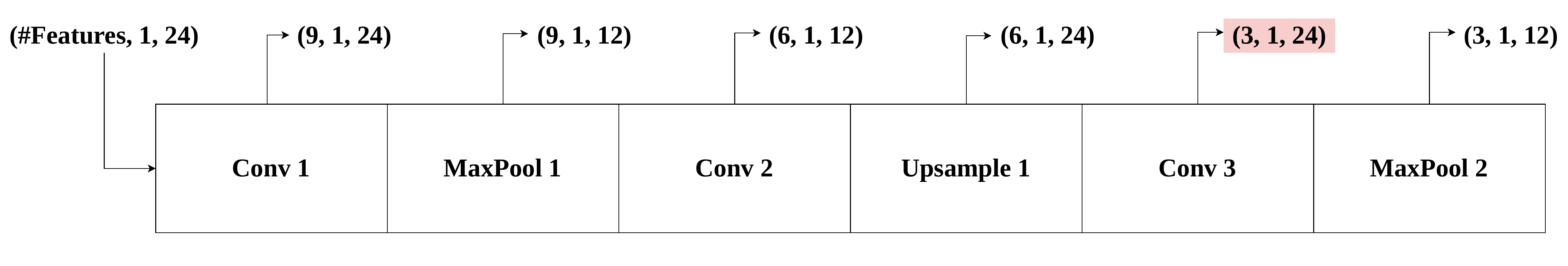}

		\caption{\small{
				{\bf Data dimension when passing to the encoder.} The figure is the encoder part of an auto-encoder, applied non-price-like features of S\&P500 market data.}}
			
			\label{fig5}
		\end{adjustwidth}
	\end{figure}

	Different arguments for Conv1d layers result in different dimensions for data. These layers must ensure that dimension of the \texttt{code} differs only in channel size with input data. This study manipulates the stride and padding arguments of Conv1d layers to keep the data width constant. Eq~\ref{eqn:1} explains how these arguments relate to each other as following:
	\begin{eqnarray}
		\mathrm{L_{out}} = \floor*{{\dfrac{{L_{in}} + {2*padding} - {dilation*(kernel\_size-1)} - {1}}{stride}}+1},
		\label{eqn:1}
	\end{eqnarray}
	
	where \texttt{Lout} and \texttt{Lin} are the lengths of the output and input data to Conv1d layers, respectively. The \texttt{dilation} is the spacing between kernel elements and is set to one by default. If stride is set to one and padding size is calculated according to Eq~\ref{eqn:2}, 
	\begin{eqnarray}
		\mathrm{padding} = \floor*{{\dfrac{kernel \_ size - 1}{2}}},
		\label{eqn:2}
	\end{eqnarray}
	the layer does not change the data width. Fig~\ref{fig5} provides an example of how encoder layers change data dimensions.
	
	\subsubsection{Decoder}
\begin{figure}[!htb]

		\begin{adjustwidth}{-1.0in}{1.0in}

			\includegraphics[width=\textwidth]{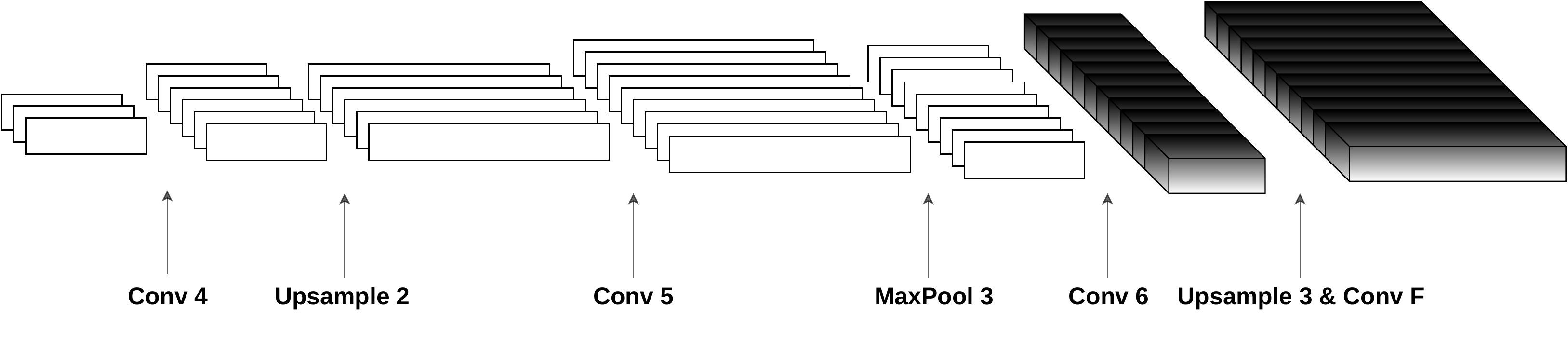}
			
		\caption{\small{
				{\bf Decoder architecture.}}}

			\label{fig6}
		\end{adjustwidth}
	\end{figure}

\begin{figure}[!htb]

	\begin{adjustwidth}{-1.0in}{1.0in}
		
		\includegraphics[width=\textwidth]{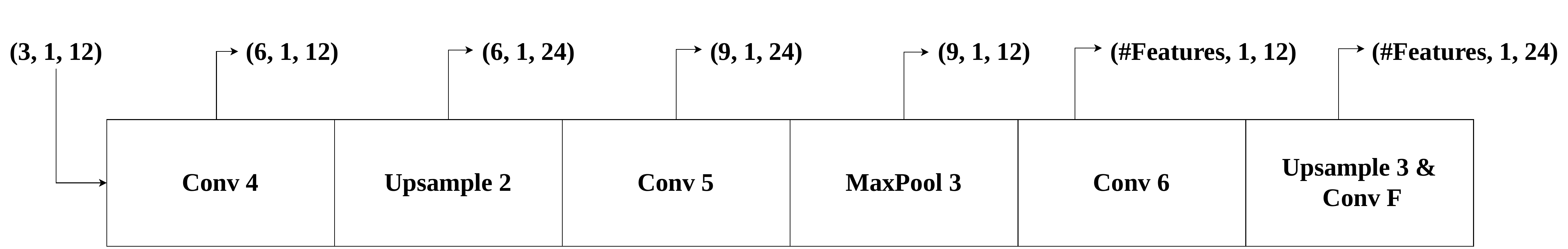}

		\caption{\small{
				{\bf Data dimension when passing to the decoder.} The figure shows the decoder part of an auto-encoder, applied non-price-like features of S\&P500 market data.}}
		
		\label{fig7}
	\end{adjustwidth}
\end{figure}

	As for the decoder part, similar layers are used with this difference that here we reconstruct the code using opposite setups for our Conv1d, max pool, and up sampling layers. Like the encoder part, Conv1d layers of the decoder part do not change any dimension but the channel size. Max pool and up sampling layers also have the same performance as the encoder part. Fig~\ref{fig6} is the decoder architecture used in this study. The only difference is that the last layer is a sequence of up sampling and Conv1d layers. Up sampling layers have no weights to update, and during backpropagation, there is nothing associated with these layers to change. As a result, ending up with these layers, our auto-encoder is not trained. That is why Fig~\ref{fig6} considers the \texttt{Conv F} right after the up sampling layer. Fig~\ref{fig7} provides an example of how decoder layers change data dimensions.

	\subsection{Overview of data preparation part}
\begin{figure}[!htb]

		\begin{adjustwidth}{-1.0in}{2.7in}

			\includegraphics[width=\textwidth]{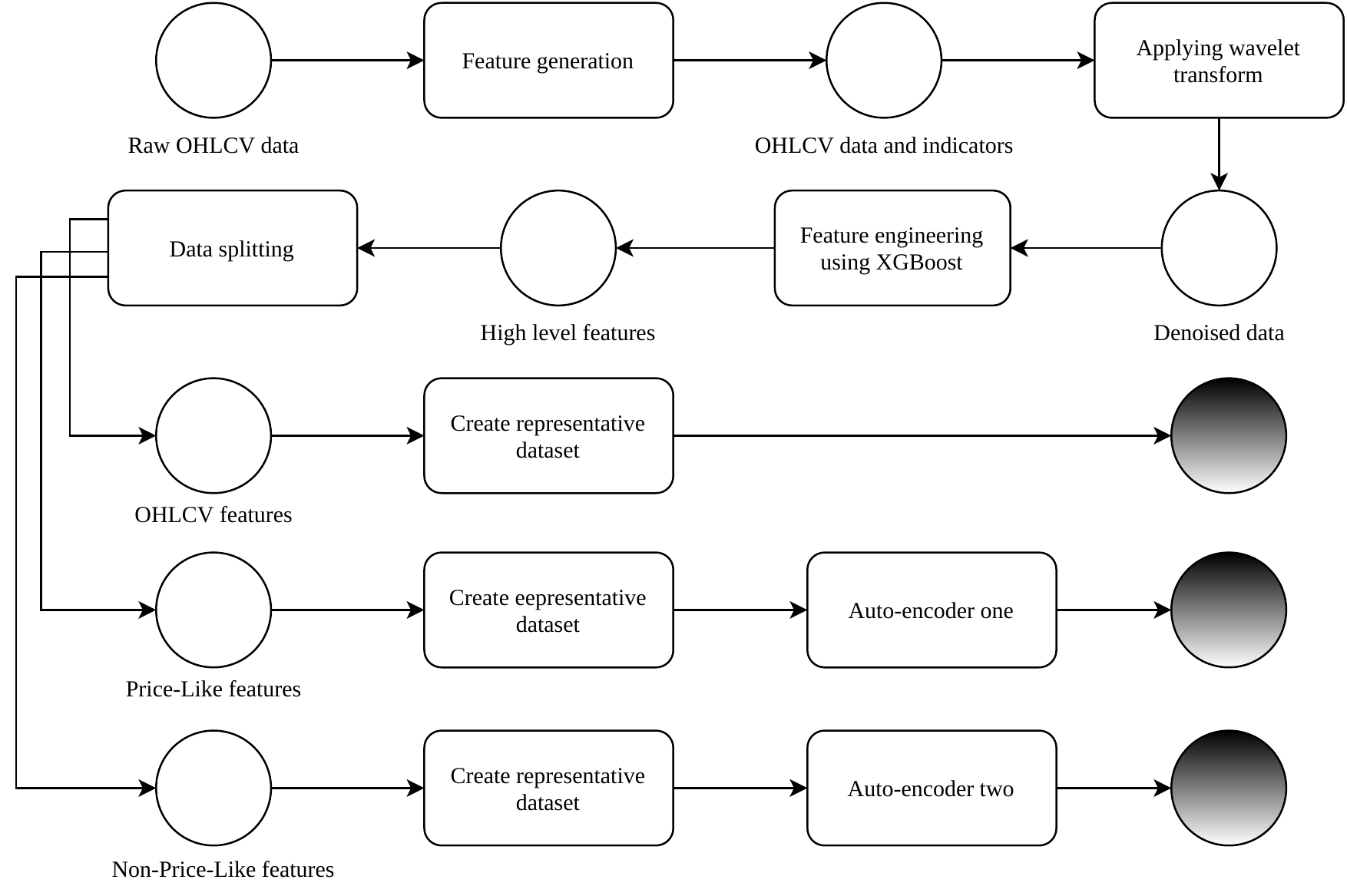}

		\caption{\small{
				{\bf Data preparation part overview.} The black-colored data is the prepared data that is ready to be fed to our main model.}}
			\label{fig8}
		\end{adjustwidth}
	\end{figure}

	To sum up the first part of this study, data preparation, Fig~\ref{fig8} provides a schematic overview of the sequence of steps being taken place till now. In the first part, to prevent confusion between the auto-encoders learning process and the primary model's learning process, this research did not explain how to label our training instances. However, in the second part it is reviewed in detail.
	
	\section{Modeling}
	The preparation of data in an accurate manner is important, but putting together a reliable market forecast network is equally important. Traders may have access to too many data, but still cannot decide whether to buy or sell. This part ensures that there is a deep neural network in charge of forecasting the price movement direction for a one-step-ahead closing price.

Similarly to the data preparation part, this part uses intuitions culled from financial markets to develop the model. Firstly, it explains how to label training instances. Once done, it constructs the body of the deep network in accordance with intuitive reasons.
	
	\subsection{Labeling}
	This step follows a straightforward labeling method. To label the first training instance, we compare the closing price of the  \texttt{window-th} candle to the closing price of the \texttt{(window+1)-th} candle. If the price went up, the label +1, and in the case, that price went down the label -1 is assigned to this instance. Eq~\ref{eqn:3},
	\begin{eqnarray}
		\begin{gathered}
			\mathcal{L}_{n} = \mathcal{F}(\mathcal{P}_{n*window}, \mathcal{P}_{n*window+1}), \\[0.4cm]
			\text{and} \\[0.2cm]
			\mathcal{F}(p, p') = Sign(\dfrac{p' - p}{p'}), \\ 
		\end{gathered}
		\label{eqn:3}
	\end{eqnarray}
	is the general formula to label training samples,
	where \texttt{${\mathcal{L}}_n \in \{-1, +1\}$}  is the label assigned to the \texttt{n-th} training instance and \texttt{${\mathcal{P}}_i$} corresponds to the closing price of the \texttt{i-th} candle.
	
	\subsection{Network architecture}
	This section introduces and designs a deep network through which binary classification tasks is performed. The network is divided into three elements. The first element includes three separate lines of convolutional layers (Conv element), each being applied on OHLCV, price-like, and non-price-like features. The next element is a many-to-one LSTM unit (LSTM element), which detects and learns the sequential patterns in data. Finally, two fully-connected layers (FC element) are considered to binary-classify the data instances.
	
	\subsubsection{Conv element}
	Before drilling down to the detailed construction of convolutional layers, it is worth reviewing why we should consider this element. We are required to design such an element due to the concept of time frame in financial markets. It is important to become familiar with the tick data on financial markets to better understand time frames. A tick is a measure of the minimum upward or downward movement in the price of a security. Each candle may contain thousands of ticks. Their corresponding quantity and price generally report ticks. Despite we can access the tick data, price charts avoid visualizing them because tick data may exceed a hundred thousand for one day. Instead, they summarize a high number of tick data to one candle. This is done by determining equal intervals and aggregating all ticks within the interval to calculate open, high, low, and close prices. This time interval is the time frame that we referred to previously. For example, when we set the time frame to one-hour, all tick data in this interval is used to form its corresponding candle.
	
	Most of the platforms which provide us analyzing tools allow us to change this time frame. It could range from one-minute to one-month. The intuition is that traders do not suffice on just one time frame to open a position. They may analyze BTC/USDT charts on different time frames and try to find a pattern. The most important thing is that after detecting a pattern, they try to give confirmation from low-frequency time frames like daily time frames. They may find a good-looking channel price for BTC/USDT in the time frame of four-hour inside which price is fluctuating between upper and lower lines. Suppose the price breaks out of the upper line, and breaks its resistance, and it seems an excellent time to open a long position in the future market or a buy position in the spot market. However, they do not open any position until getting a strong confirmation that this breakout is not fake. One way to give the confirmation is to change the time frame and observe the price position in the daily time frame. Generally, financial events such as breaking out a channel price or pulling back to the broken resistance or support lines are more valid on low-frequency time frames like daily or weekly time frames.
	
\begin{figure}[!htb]

		\begin{adjustwidth}{-1.0in}{2.7in}

			\includegraphics[width=\textwidth]{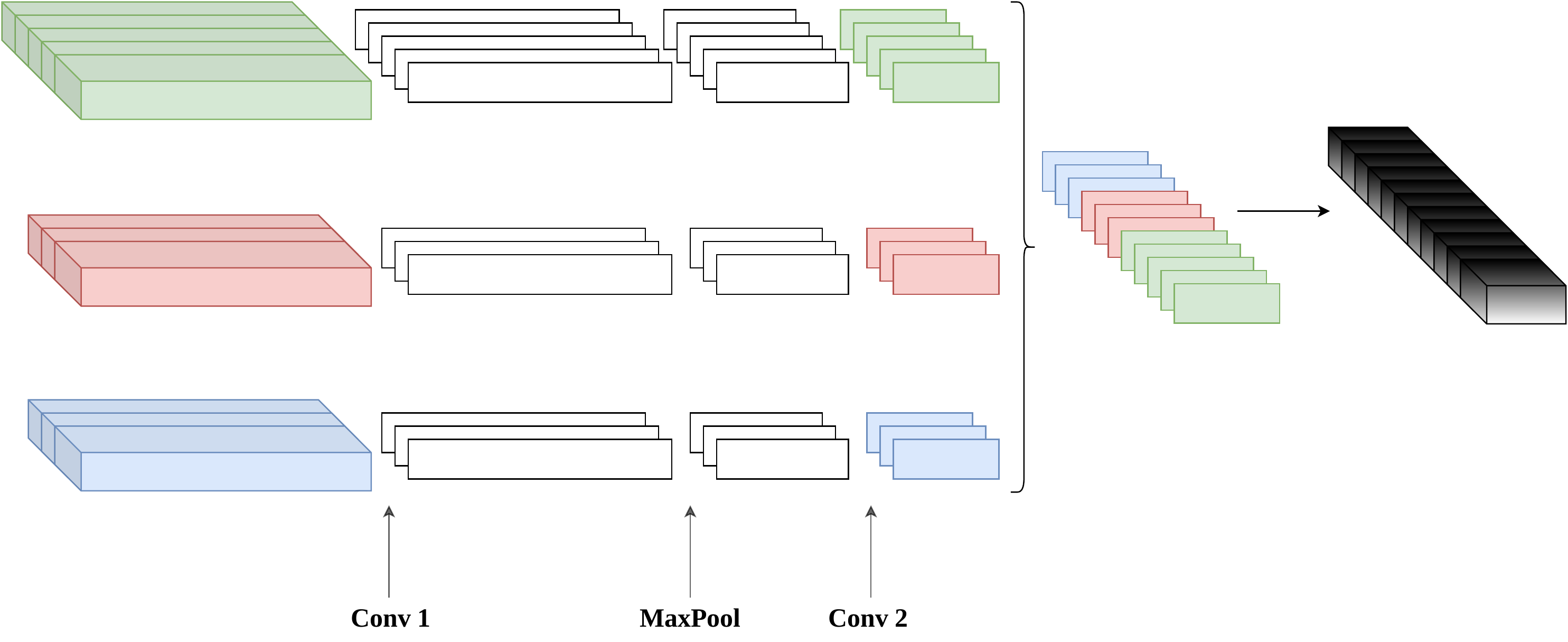}
		
		\caption{\small{
				{\bf Conv element - the first element of the model.} The green, red and blue colored data are OHLCV, price-like and non-price-like data respectively.}}
			\label{fig9}
		\end{adjustwidth}
	\end{figure}

\begin{figure}[!htb]

		\begin{adjustwidth}{-1.0in}{2.7in}

			\includegraphics[width=\textwidth]{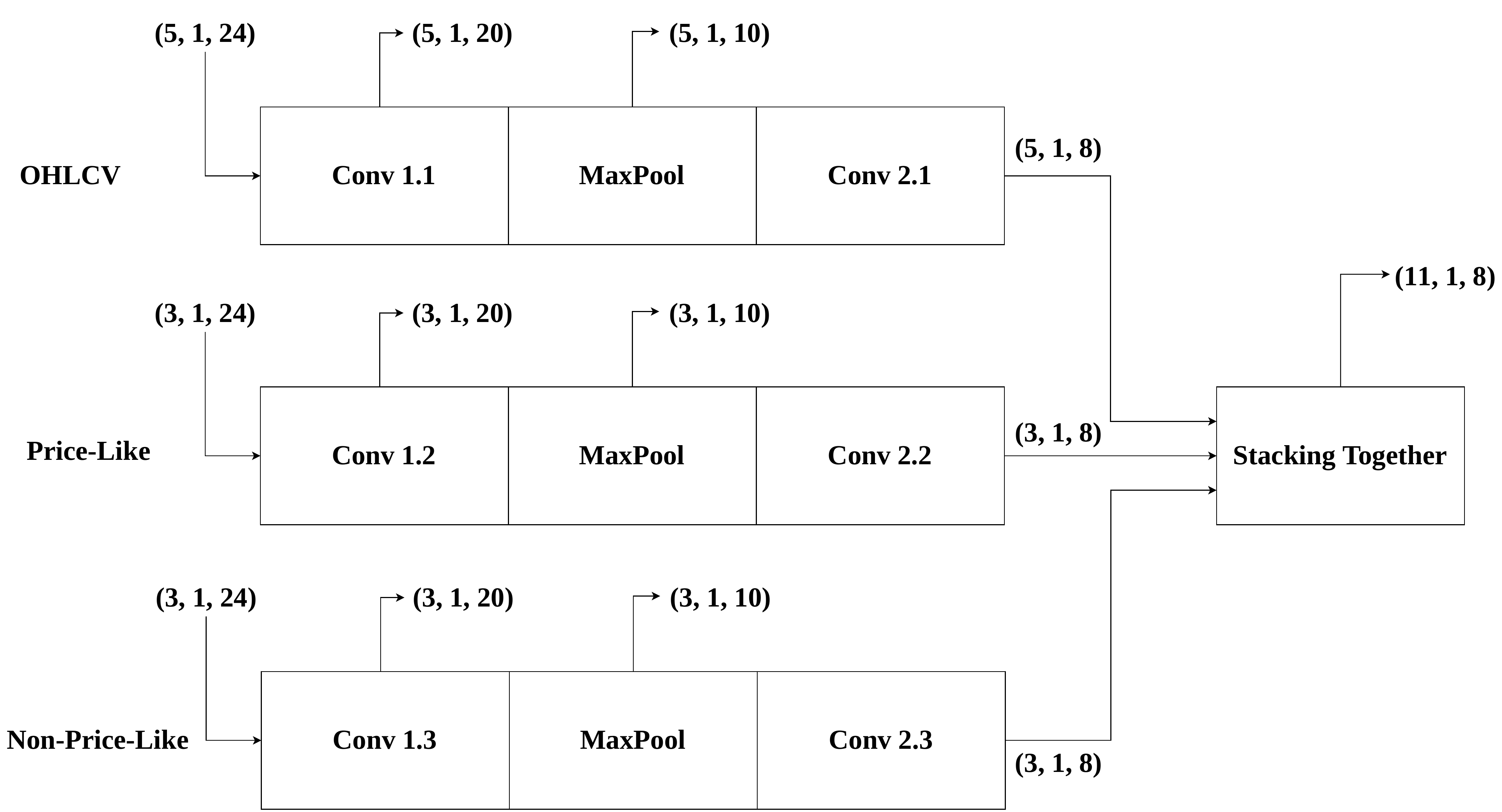}
		
		\caption{\small{
				{\bf Data dimension when passing to the Conv element.} The figure shows a Conv element, applied on training instances of S\&P500 market data. Reducing the width from 24 to 8 could be considered like changing the time frame from 8h to daily.}}
			\label{fig10}
		\end{adjustwidth}
	\end{figure}

	This section considers convolutional layers whose duty is to reduce the width size of every training instance. Recall that the width size is the number of consecutive candles through which we want to analyze the price path. Reducing this number is exactly like using low-frequency time frames. This paper designs three separate layers containing one max pool and two one-dimensional convolutional layers for OHLCV, price-like and non-price-like data. Fig~\ref{fig9} shows the detailed construction of these lines. It also illustrates that we finally stack outputs together for the LSTM element. Fig~\ref{fig10} also shows how these layers and lines are changing the data dimension.
	
	\subsubsection{LSTM element}
	In financial markets, the price path is much more important than the current absolute price. The reason is that events on previous candles may affect price behavior on the next candles. When determining static or dynamic support and resistance lines, traders use previous candles since they believe the information coming from previous candles can be used to analyze the price fluctuations. RNNs~\cite{rout2017forecasting, tino2001financial} are designed precisely to address such issues in which the information from the past is also essential for current prediction. Hidden states are in charge of transforming this information.
	
	One of the disadvantages associated with RNNs is gradient vanishing when backpropagating through time~\cite{hochreiter1998vanishing, pascanu2013difficulty}. It causes long-term dependencies are not detected. To improve the performance of RNNs to also detect long term dependencies, LSTM units are introduced~\cite{hochreiter1997long}. However, LSTM is not the only way to address memory problem. Gated recurrent units (GRUs), the simpler version of LSTM units~\cite{chung2014empirical}, address the same problem either. The idea behind either LSTM or GRU is to consider a memory cell that determines how much information should be memorized and transformed to the next units. LSTM with three gates named forget gate, update gate, and output gate, and GRU with one gate called update gate, handle this memory transformation.
	
Fig~\ref{fig10} indicates that the data sequence is not long enough to cause gradient vanishing. However, this paper prefers to use LSTM units to detect chronological patterns. It undergoes a more computational process but ensures that required information is passed to the next units as much as possible. 
	
\begin{figure}[!htb]

		\begin{adjustwidth}{-1.0in}{2.7in}
		
			\includegraphics[width=\textwidth]{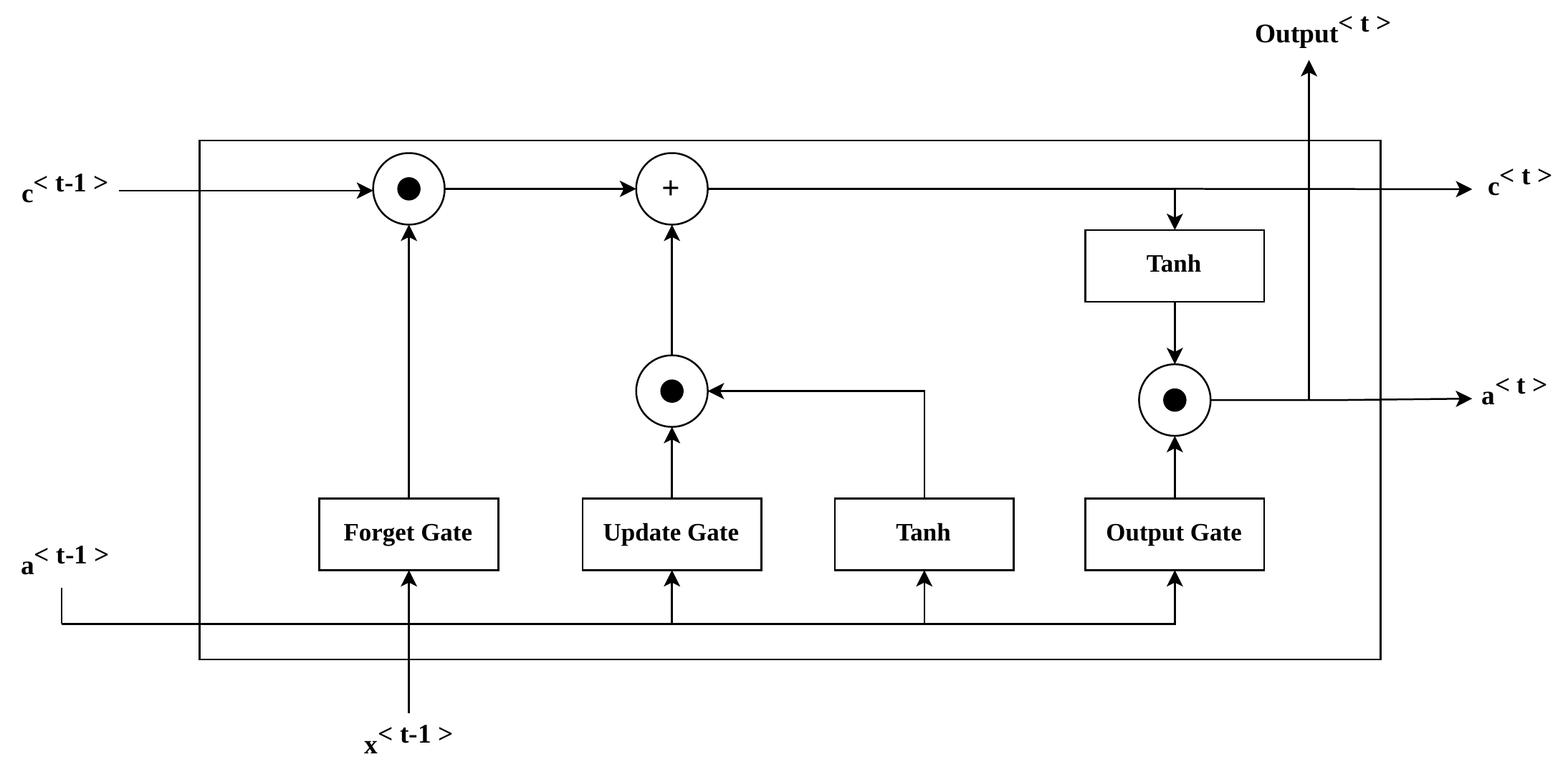}

		\caption{\small{
				{\bf Single LSTM unit architecture.}}}
			\label{fig11}
		\end{adjustwidth}
	\end{figure}

\begin{figure}[!htb]

		\begin{adjustwidth}{-1.0in}{2.7in}
		
			\includegraphics[width=\textwidth]{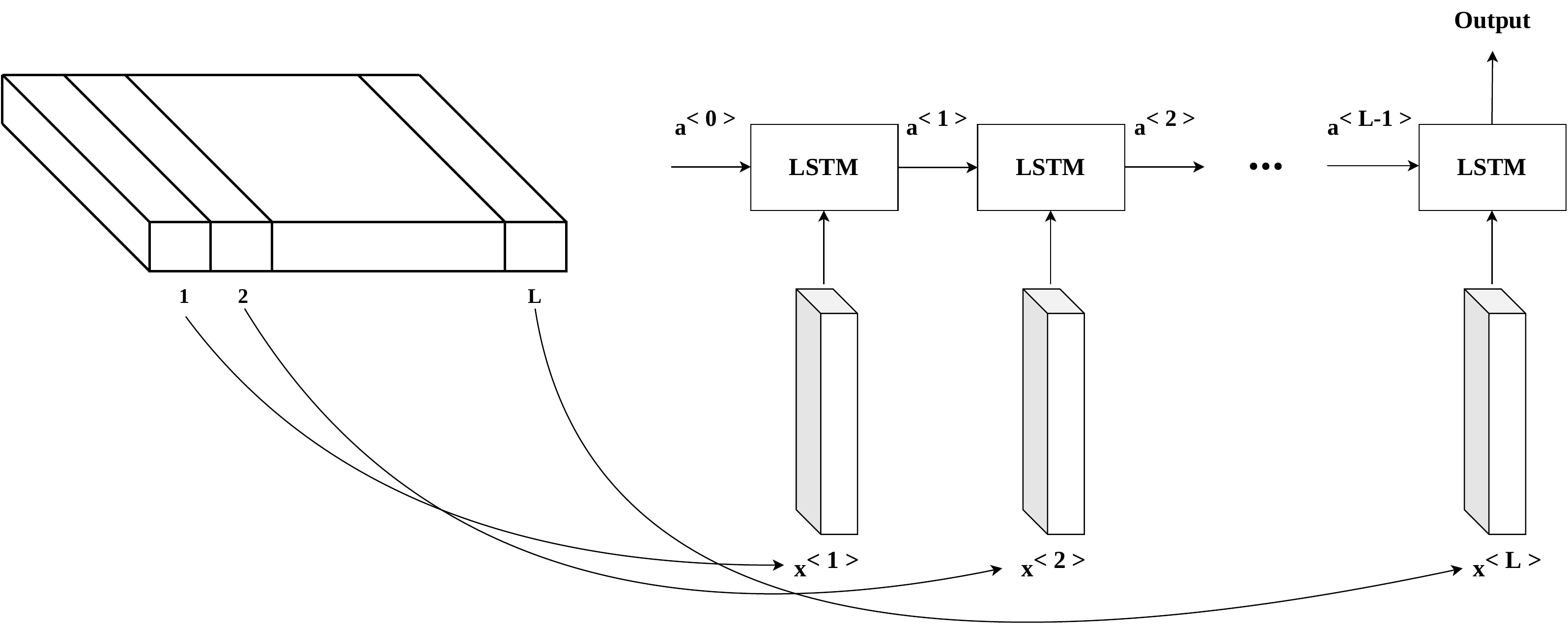}

		\caption{\small{
				{\bf LSTM element - the second element of the model.} {${x}^{<t>}, {a}^{<t>}$} are the data sequence and hidden states respectively.}}
			\label{fig12}
		\end{adjustwidth}
	\end{figure}

	Compared to RNNs, LSTM takes advantage of a memory cell $c^{<t>}$ to transform information. $c^{<t>}$ is calculated by weighting to the previous memory cell $c^{<t-1>}$ and a candidate cell $\tilde{c}^{<t>}$ which is related to the current LSTM unit. The update and forget gates handle the weighting task. They specify how much memory from previous units must be transformed and how much information from the current unit must be considered. Having $c^{<t>}$ calculated, the output gate is used to calculate the current hidden state  $a^{<t>}$. Finally, $a^{<t>}$ is used to make a prediction. As LSTM units are consecutive, $a^{<t>}$ and $c^{<t>}$ are transformed to the next LSTM unit, and this process continues until the last unit. The following equations describe the order in which each of the above elements is calculated:
	\begin{eqnarray}
		{\tilde{c}^{<t>}} = \mathrm{tanh}(W_{c}[a^{<t-1>}, x^{<t>}]+b_{c}),
		\label{eqn:4}
	\end{eqnarray}
	\begin{eqnarray}
		{ \Gamma_u} = \mathrm{\sigma}(W_{u}[a^{<t-1>}, x^{<t>}]+b_{u}),
		\label{eqn:5}
	\end{eqnarray}
	\begin{eqnarray}
		{ \Gamma_f} = \mathrm{\sigma}(W_{f}[a^{<t-1>}, x^{<t>}]+b_{f}),
		\label{eqn:6}
	\end{eqnarray}
	\begin{eqnarray}
		{ \Gamma_o} = \mathrm{\sigma}(W_{o}[a^{<t-1>}, x^{<t>}]+b_{o}),
		\label{eqn:7}
	\end{eqnarray}
	where \texttt{$W$}s and \texttt{$b$}s are weight matrices which are updated through backpropagation step and biases respectively. \texttt{$\Gamma_u$}, \texttt{$\Gamma_f$} and  \texttt{$\Gamma_o$} are update gate, forget gate and output gate respectively. \texttt{$\sigma$} is also the sigmoid activation function. Using these gates along with candidate cell $\tilde{c}^{<t>}$, Eq~\ref{eqn:8} and Eq~\ref{eqn:9} calculate $c^{<t>}$ and $a^{<t>}$ to be passed through the rest of LSTM units as following:
	\begin{eqnarray}
		{{c}^{<t>}} = \mathrm{\Gamma_u}\circ{\tilde{c}^{<t>}} + \mathrm{\Gamma_f}\circ{{c}^{<t-1>}}
		\label{eqn:8},
	\end{eqnarray}
	\begin{eqnarray}
		{{a}^{<t>}} = \mathrm{\Gamma_o}\circ\mathrm{tanh}(c^{<t>}).
		\label{eqn:9}
	\end{eqnarray}
	In each unit, the corresponding value of the forget gate determines how much information we are better to forget. If it is close to 1, the previous memory cell contains valuable information from the past and should be transformed to the current memory cell. If it is close to 0, it means that the major part of the current memory cell is calculated by the candidate cell.
The \texttt{$\Gamma_u$}, \texttt{$\Gamma_f$} and \texttt{$\Gamma_o$} are not constant for all LSTM units. They are vectors containing the sigmoid output for each LSTM unit. It is worth mentioning that gates, hidden states, memory cells, and candidate cells are all vectors, so Eq~\ref{eqn:8} and Eq~\ref{eqn:9} use element-wise products between these vectors. As the last step, if we wish to predict the current LSTM unit, Eq~\ref{eqn:10},
	\begin{eqnarray}
		{\hat{y}^{<t>}} = \mathrm{g}(W_{y}a^{<t>}+b_{y}),
		\label{eqn:10}
	\end{eqnarray}
	is employed to return the output where \texttt{g} is an activation function. Since LSTM units are not generally used to directly binary-classify the data instances, most of the time, like in this paper, \texttt{g} is a softmax activation function. Fig~\ref{fig11} illustrates a schematic overview of a classic LSTM unit. There are other kinds of LSTM units such as \texttt{peephole connections} which apart from $a^{<t-1>}$ and $x^{<t>}$ considers $c^{<t-1>}$ to calculate gates values in Eq~\ref{eqn:5}, Eq~\ref{eqn:6} and Eq~\ref{eqn:7}~\cite{gers2002learning}. This study takes advantage of a many-to-one LSTM architecture, provided in Fig~\ref{fig12}.
	
\begin{figure}[!htb]

		\begin{adjustwidth}{-1.0in}{2.7in}
			
			\includegraphics[width=\textwidth]{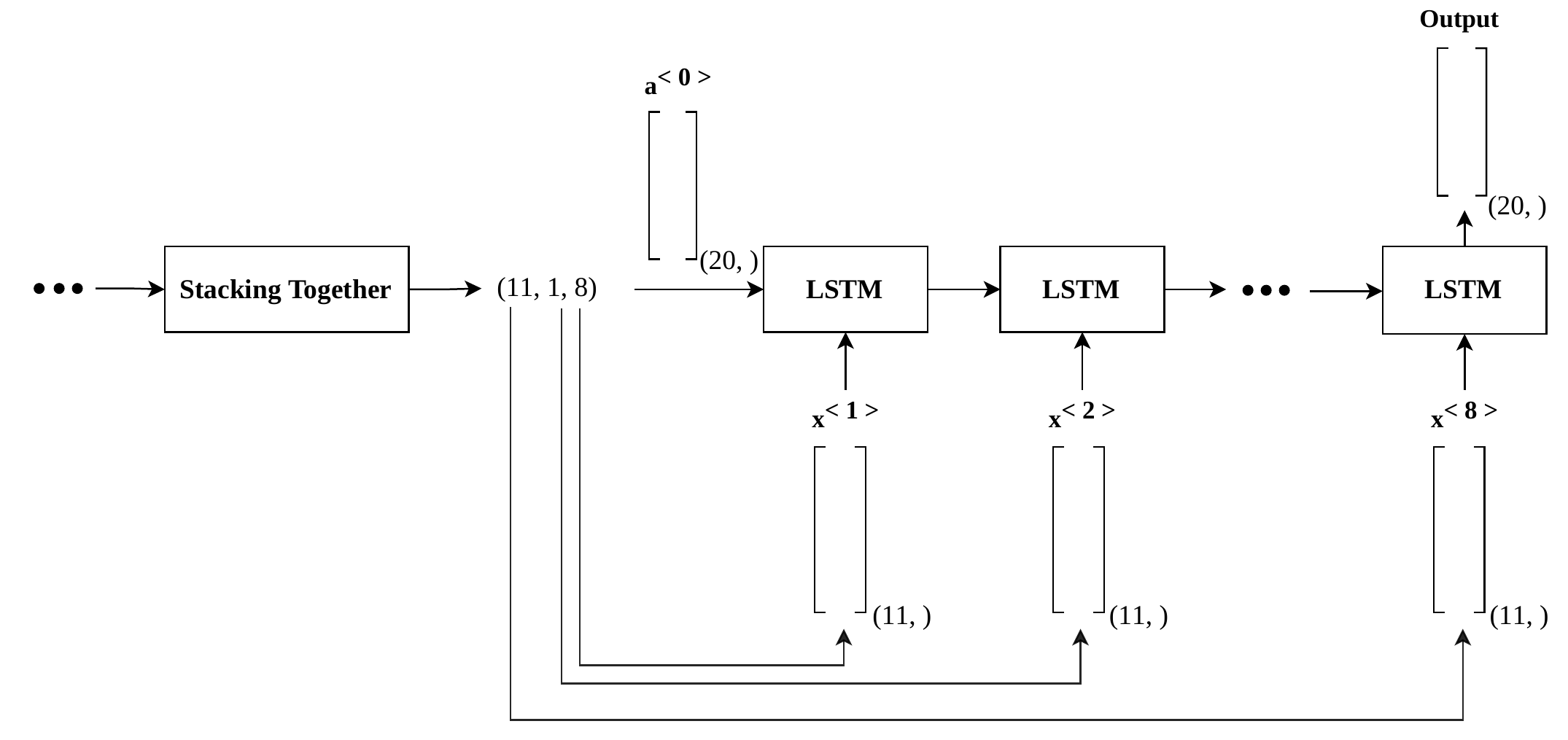}

			\caption{\small{
					{\bf Data dimension when passing to the LSTM element.} The figure shows the implementation of the LSTM element on S\&P500 market data. Output and hidden states vectors have the same length of 20. }}
				\label{fig13}
		\end{adjustwidth}
	\end{figure}

	The output length can vary according to the hidden states' dimension. This dimension is a hyperparameter to set. This paper designs LSTM units which require the output to be the same length as hidden states' length. Fig~\ref{fig13} illustrates how this process happens.
	
	\subsubsection{FC element}
\begin{figure}[!htb]

		\begin{adjustwidth}{-1.0in}{1.0in}
			
			\includegraphics[width=\textwidth]{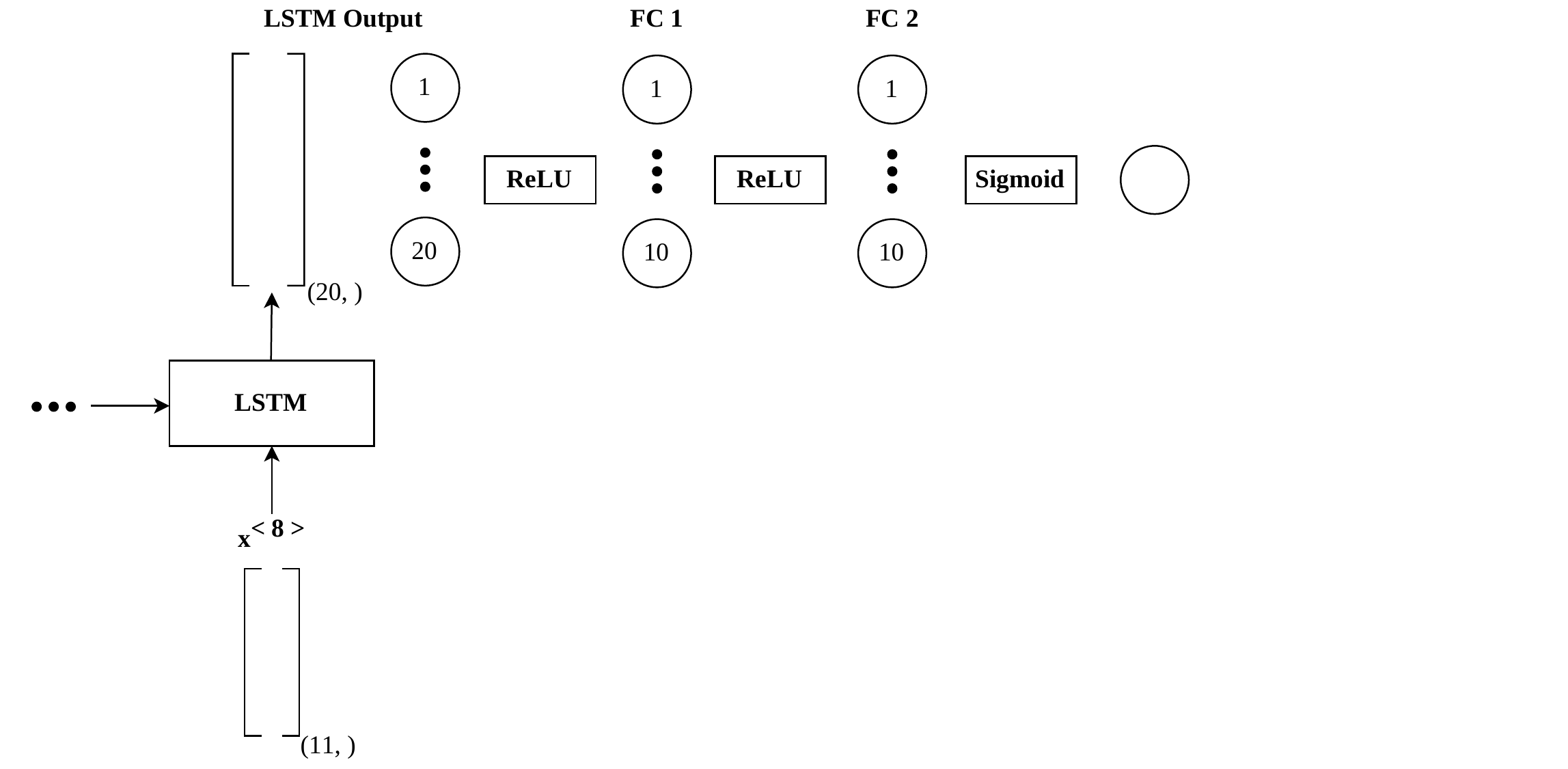}
		
		\caption{\small{
				{\bf FC element - the third element of the model.} The figure shows the implementation of the FC element on S\&P500 market data.}}
			\label{fig14}
		\end{adjustwidth}
	\end{figure}

	As the last element of the model, fully-connected layers binary-classify the LSTM output vector. Despite its name, the layers used in this section are not really fully-connected since dropout layers are used to protect the model from over-fitting~\cite{srivastava2014dropout}. Two fully-connected layers are used to determine whether the next close price goes up or down. Fig~\ref{fig14} illustrates applying these layers to the LSTM output on actual data. 

	\subsection{Model quality check}
	When it comes to financial time series data, the model is occasionally not trained at all. When back-propagating, the loss would decrease, but it is not significant to claim the model is trained. In other words, the model loss converges too soon. For example, you may consider 1000 epochs to train your model, but the model loss decreases until the 200th epoch, and for the rest of the epochs, there is no significant reduction in model loss. This section will discuss how the training process can be checked.

In the financial time series, both model over-fitting and under-fitting are issues. The concern of over-fitting is abated by training the model on the high number of data and using denoisers and dropout layers in the network. This part mainly talks about an index by which we can make sure our model is not under-fitted and can detect patterns. The proposed index to quality check the training process is obtained by reviewing the loss function. Eq~\ref{eqn:11}, 
	\begin{eqnarray}
		{L(y, \hat{y})} = \mathrm{-\dfrac{1}{m}}(\sum^{m}_{i=1} \ y_i \log \hat{y}_i + (1-y_i)\log(1-\hat{y}_i)),
		\label{eqn:11}
	\end{eqnarray}
	is a binary cross-entropy loss function widely used in binary classification tasks where \texttt{${y}_i \in \{0, 1\}$}, \texttt{${\hat{y}}_i \in (0, 1)$} and \texttt{m} are the actual label of \texttt{i-th} data instance, predicted label of \texttt{i-th} data instance and number of data instances respectively. Through the backpropagation, the loss function is supposed to decrease. The equation says, on average, how wrong the model is classifying data instances.

\begin{figure}[!htb]

		\begin{adjustwidth}{-1.0in}{2.7in}

			\includegraphics[width=\textwidth]{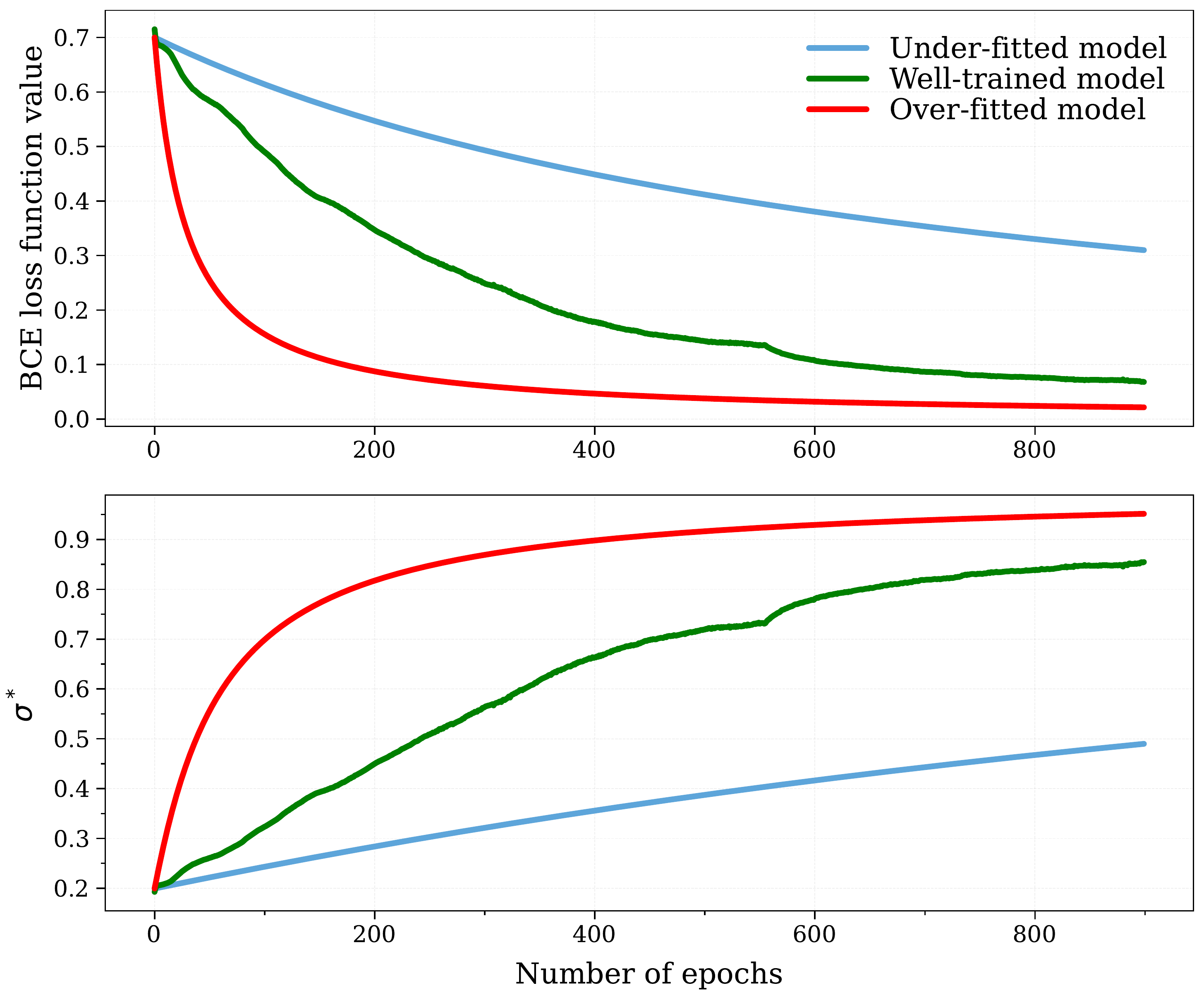}
			
			\caption{\small{
					{\bf Loss function value role to measure training quality.} The figure shows the loss function and $\sigma^*$ values for three categories of models. The green one belongs to a model trained on BTC/USDT data (time frame of four-hour). The converged loss function is near 0.07, which means that if we consider all of the data instances as a single instance with label 1, the model correctly predicts this instance with the Sigmoid value of 0.85.}}
				\label{fig15}
		\end{adjustwidth}
	\end{figure}

This paper firstly suggests grabbing the value of the converged loss function and then calculating its corresponding sigmoid value. It is obtained by using Eq~\ref{eqn:12},
	\begin{eqnarray}
		\sigma^* = 10 ^ {- L_c},
		\label{eqn:12}
	\end{eqnarray}
	where \texttt{$\sigma^*$} is the corresponding sigmoid value of the converged loss function \texttt{$L_c$}. The \texttt{$\sigma^*$} represents what sigmoid value, on average, the model would have if we suppose the whole data as a single data instance with an actual label of 1. According to Eq~\ref{eqn:13}, by putting a threshold on \texttt{$\sigma^*$}, we can figure out whether the model is trained properly:
	\begin{eqnarray}
		\mathrm{Training \ quality}=
		\begin{cases}
			\text{Well-trained}: & \text{if}  \ \ \sigma^* \geq \zeta \\
			\text{Under-fitted}: & \text{if} \ \ \sigma^* < \zeta
		\end{cases} ,
		\label{eqn:13}
	\end{eqnarray}
	where \texttt{$\zeta$} is an optional threshold for the \texttt{$\sigma^*$}. This paper uses a value of between 0.8 and 0.85 for \texttt{$\zeta$}, Taking advantage of grid-searching. Suppose it is set to 0.8, equivalent to the loss function of 0.07 according to Eq~\ref{eqn:12}. In this case, it means that if we consider all data instances as a single data instance with label 1, the model classifies this single data instance with the sigmoid value of 0.8, which is confident enough to claim the model is trained. On the other hand, if the model loss function converges to 0.5 too soon and does not change anymore, it means that if we consider all data instances as a single data instance with label 1, the model classifies this single data instance with the sigmoid value of 0.31 which is not confident at all and is a sign of model under-fitting. Forcing the model to pass higher values of $\zeta$ results in over-fitting. Fig~\ref{fig15} illustrates a plotted loss function for a healthy trained model, which is neither over-fitted nor under-fitted.
	
\begin{figure}[!htb]
		\begin{adjustwidth}{-1.0in}{2.7in}
		
			\includegraphics[width=\textwidth]{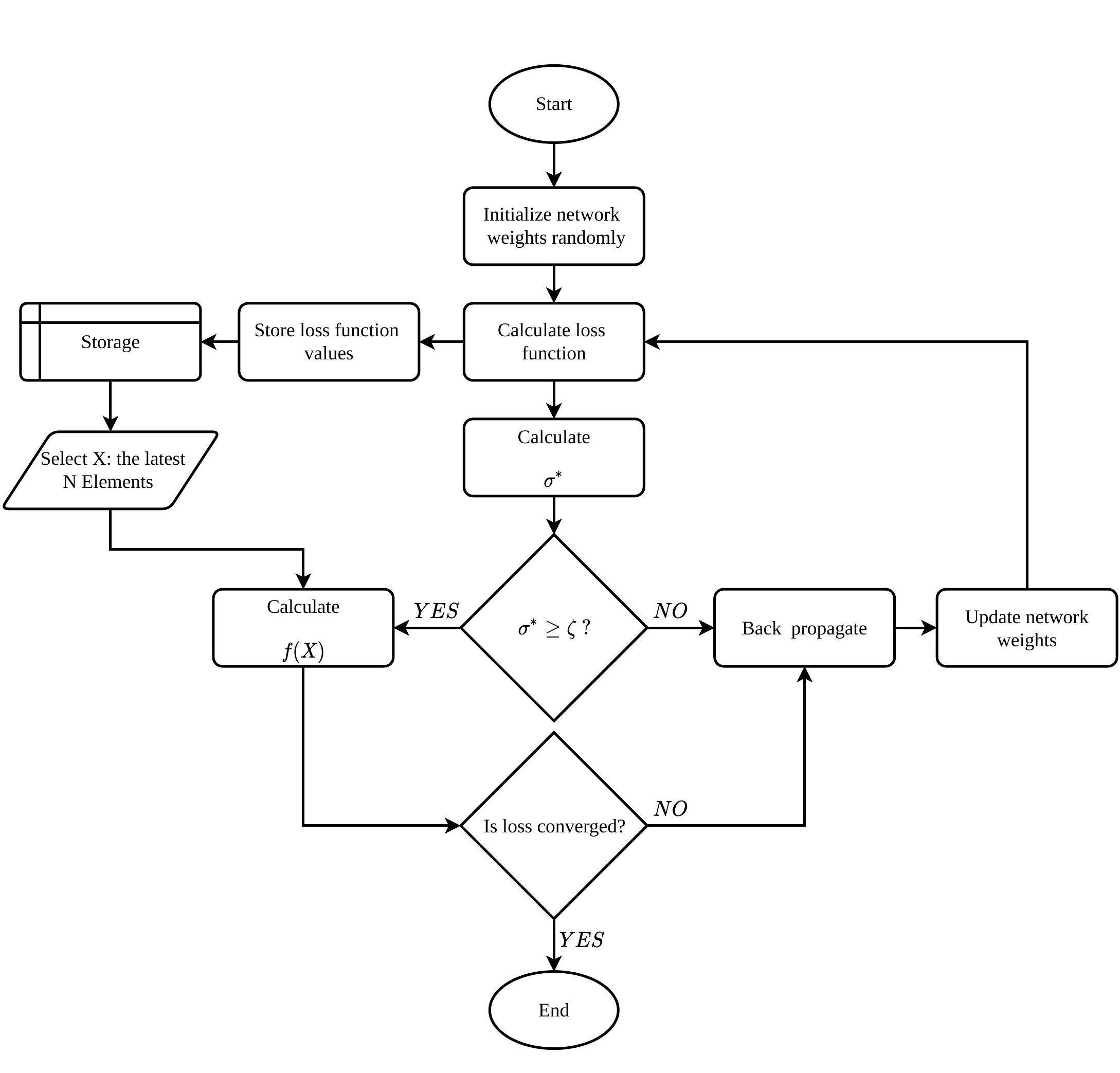}

		\caption{\small{
				{\bf Model training flow chart.} This paper calculates the $\sigma^*$ for each epoch. If it was equal or greater than $\zeta$, the latest $N$ loss function values($X$) are retrieved from the storage. Then the function $f$ decides whether the loss has converged. If so, it breaks the loop and returns the trained model. Function $f$ considers the first element of $X$ as the primary point and then calculates the slope of $N-1$ lines, starting from the primary point and ending at other elements of $X$. If the average of these slopes is less than a threshold, it means that the loss is converged. The suitable threshold is a negative value close to zero.}}
			\label{fig16}
		\end{adjustwidth}
	\end{figure}

	Studies typically use for-loops in the code when training the network to make the model iterate for a certain number of epochs. However, this research uses a while-loop instead. Whenever the converged loss resulted in a $\sigma^*$ greater than $\zeta$, it breaks the while-loop. Fig~\ref{fig16} illustrates the model training process.

	\section{Trading strategy}
	Through data preparation and modeling parts, we classified our data instances to predict the next closing price. However, it does not mean the currently trained model is ready to act on financial markets. When using Artificial intelligence (AI), we should carefully map the model outputs to our needs; otherwise, the outputs are not helpful in our real-world problems. Every unrealistic assumption to virtually test our model's performance does not help us to detect weaknesses. While we expect a sixty-percent accurate model to be profitable, it may make us lose the margin. This part aims to introduce a trading strategy whose essential duty is to ignore all unrealistic assumptions by selecting informative indicators to evaluate the model.
	
	\subsection{Unrealistic assumptions}
	One of the drawbacks of using this model without further measurements is that it acts on every candle. It does not make sense since no trader acts this way. Even having a one hundred percent accurate model, acting on every candle costs a fortune for us because exchanges require the transaction fee for each position we wish to open. It is the first unrealistic assumption that must be amended.

	The second unrealistic assumption is that to evaluate the model performance, we use some indicators such as accuracy, which considers equal weights for positions. In other words, it does not matter how big the positions' profit/loss is. For instance, positions with ten percent profit are treated the same as positions with one percent profit. Furthermore, accuracy does not care about the positions' margin. No matter how much money we specify to one position, it just checks whether the price direction was truly predicted or not. It is not close to mind since traders do not necessarily divide their margin equally for their positions. It is also important for them how much profit they earn from each position. They may open eight wrong positions but still be satisfied since their other two correct positions give them much more profit than the amount of loss coming from their wrong positions. That is why accuracy is not a solid and creditable index to evaluate the model performance.
	
	\subsubsection{Acting on every single candle}
	To deal with the first unrealistic assumption, acting on every single candle, this research recommends setting a threshold for the output coming from the last sigmoid function prior to binary-classifying data. The sigmoid production is between 0 and 1. A common approach to classify the data is to consider label 0 for those instances with sigmoid output lower than 0.5 and label 1 for instances with sigmoid output equal or greater than 0.5. By changing this threshold, we can sift predictions with higher confidence.

\begin{figure}[!htb]

		\begin{adjustwidth}{-1.0in}{2.7in}

			\includegraphics[width=\textwidth]{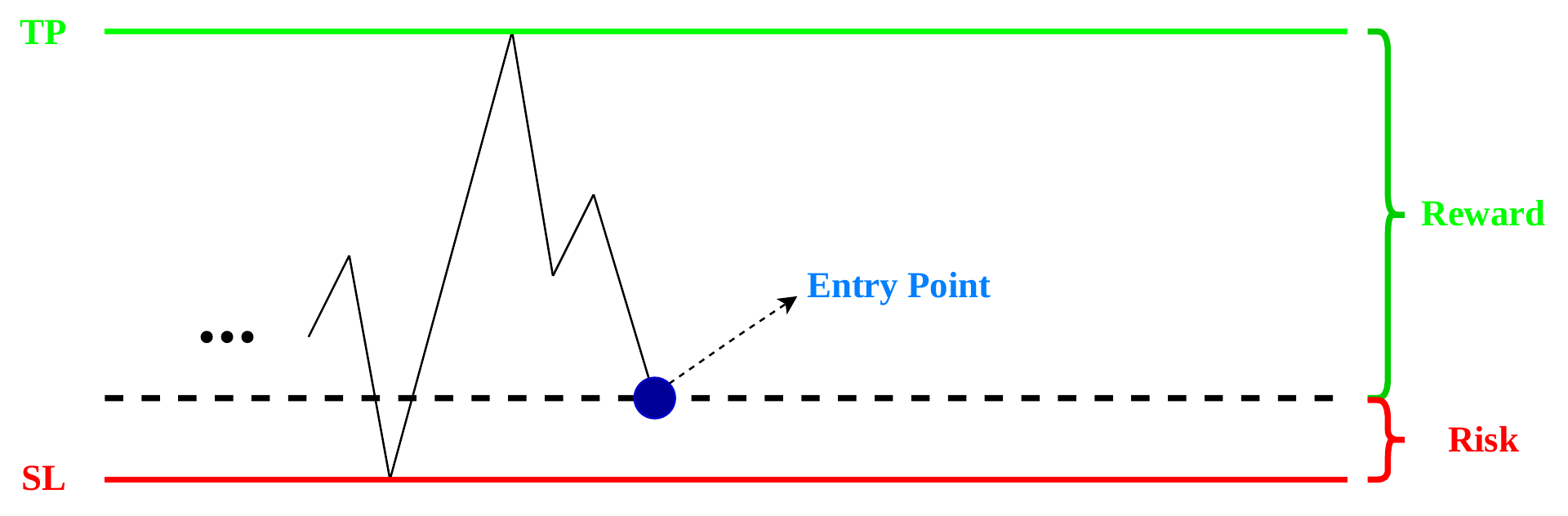}

		\caption{\small{
				{\bf Stop loss and take profit prices.} Setting SL and TP prices is a form of art and traders use different strategies to do that. This figure uses the highest and lowest prices as TP and SL since these prices can play a support and resistance lines for this snapshot.}}
			\label{fig17}
		\end{adjustwidth}
	\end{figure}

Traders may detect good entry points to open a position, but they do not risk opening all of them. They tend to measure an index named \texttt{Risk/Reward} firstly. The less this index~\cite{gilli2011risk}, the more attractive the position  appears for traders. The inverted version of this index says how big the estimated reward (profit) is compared to the potential loss we undergo if we open a position. It is generally calculated by Stop loss (SL) and Take profit (TP) lines which traders set before opening a position. The SL is a price on which traders no longer wish to undergo loss. Similarly, if the price moves according to their expectation, TP is defined as an earlier price on which traders wish to save their profit. The Risk/Reward index is nothing but the absolute difference between entry point and SL price divided by the absolute difference between TP and entry point price as indicated on Eq~\ref{eqn:14}. Fig~\ref{fig17} shows how this index is calculated according to SL and TP prices.

	\begin{eqnarray}
		\mathrm{Risk/Reward} = \frac{|E - SL|}{|TP - E|}
		\label{eqn:14},
	\end{eqnarray}
	where \texttt{E} is an entry point price. Traders want this index as little as possible. However, each of them has the desired value for this index at which they enter a position. The most common value for this index is $\frac{1}{3}$ among traders.

The threshold for the sigmoid function can be determined according to the Risk/Reward index. This paper refers to this index as $\theta$ for ease of pointing. Eq~\ref{eqn:15},
	\begin{eqnarray}
\begin{gathered}
\mathrm{Upper} = \frac{1}{1+\theta} \\
\mathrm{Lower} = \frac{\theta}{1+\theta}
\end{gathered},
\label{eqn:15}
\end{eqnarray} defines two thresholds for sigmoid output, named upper and lower thresholds, respectively. Data instances with a sigmoid value greater than the upper threshold or less than the lower threshold are selected as entry points. This way, we do not act on those candles with a sigmoid output of between upper and lower thresholds. Eq~\ref{eqn:16},
	\begin{eqnarray}
		\mathrm{Signal}=
		\begin{cases}
			\text{Long/Buy}: & \text{if}  \ \ \sigma_t \geq Upper \\
			\text{Short/Sell}: & \text{if} \ \ \sigma_t \leq Lower
		\end{cases} ,
		\label{eqn:16}
	\end{eqnarray}
	 describes when this strategy opens either a long or short position where \texttt{$\sigma_t$} is the value of the sigmoid function corresponding to the \texttt{t-th} data instance. Depending on the value of $\theta$, this strategy ignores a part of positions whose confidence is not adequate. Consequently, the problem of acting on each candle is addressed.
	\subsubsection{Bet sizing}
	To address the second issue, specifying equal amounts of margin and weights for each position, this part considers a simple approach to specify positions' margins. By considering the absolute difference between the sigmoid function value and 0.5, positions' margins are calculated. For example, if the sigmoid function value is 0.1 and this number is less than the lower threshold, we specify {0.5 - 0.1 = 0.4} of our total margin for this position. By increasing the lower threshold or decreasing the upper threshold, we can allow our strategy to open much more positions, each having much less amount of margin.
	
	\subsection{Evaluate the performance}
	There are various indexes by which the performance of models is evaluated, especially in our case that the model's task is to binary-classify the data. For example, we can calculate the accuracy, recall and precision indexes.
	
	These indicators are related to the model outputs, not to trading strategy outputs. We may have an excellent trained model that is 60 percent accurate, but it makes us lose  money since positions' margins and their corresponding profit/loss may not be equal. This model may correctly predict 60 percent of instances whose corresponding margins and profits are much lower than the margins and loss of another 40 percent of instances. For example, if we correctly open 60 positions with an average initial margin of $\$$100 and an average profit of 4$\%$ and wrongly open 40 positions with an average initial margin of 150$\$$ and an average loss of 5$\%$, on the whole, we lose money. In contrast, a 45 percent accurate model would be profitable, hence accuracy is not a good indicator to determine how rich we get if we use this trained model. Although this paper reports the confusion matrix to show the model's performance, it highly recommends using absolute return to evaluate the trading strategy performance since we are making money from the trading strategy and not from the direct output of the model.
	\section*{Results}
	Selected data to train the model are four-hour data of ETH/USDT and BTC/USDT, each starting from 2017 and ending in 2021, and daily data of S\&P500 index beginning from 2001 and ending at 2021. BTC and ETH models are tested on the period of one year of data, and the S\&P500 model is tested on four years of data.
	
	This research reports two sets of results. The first one  investigates the model performance and the second and more important one reports the profitability of the model.
	
	\subsection*{Model performance}
	In order to evaluate the model performance, this part reports accuracy, recall, precision, F1 score, converged value of loss function, and its corresponding $\sigma^*$. These indexes are provided for different values of $\theta$ according to Table~\ref{table2}, Table~\ref{table3} and Table~\ref{table4}.
	
	\begin{table}[!htb]
		
		\begin{adjustwidth}{-1.0in}{2.5in}
			\caption{
				{\bf Model performance - S\&P500}}
			\includegraphics[width=\textwidth]{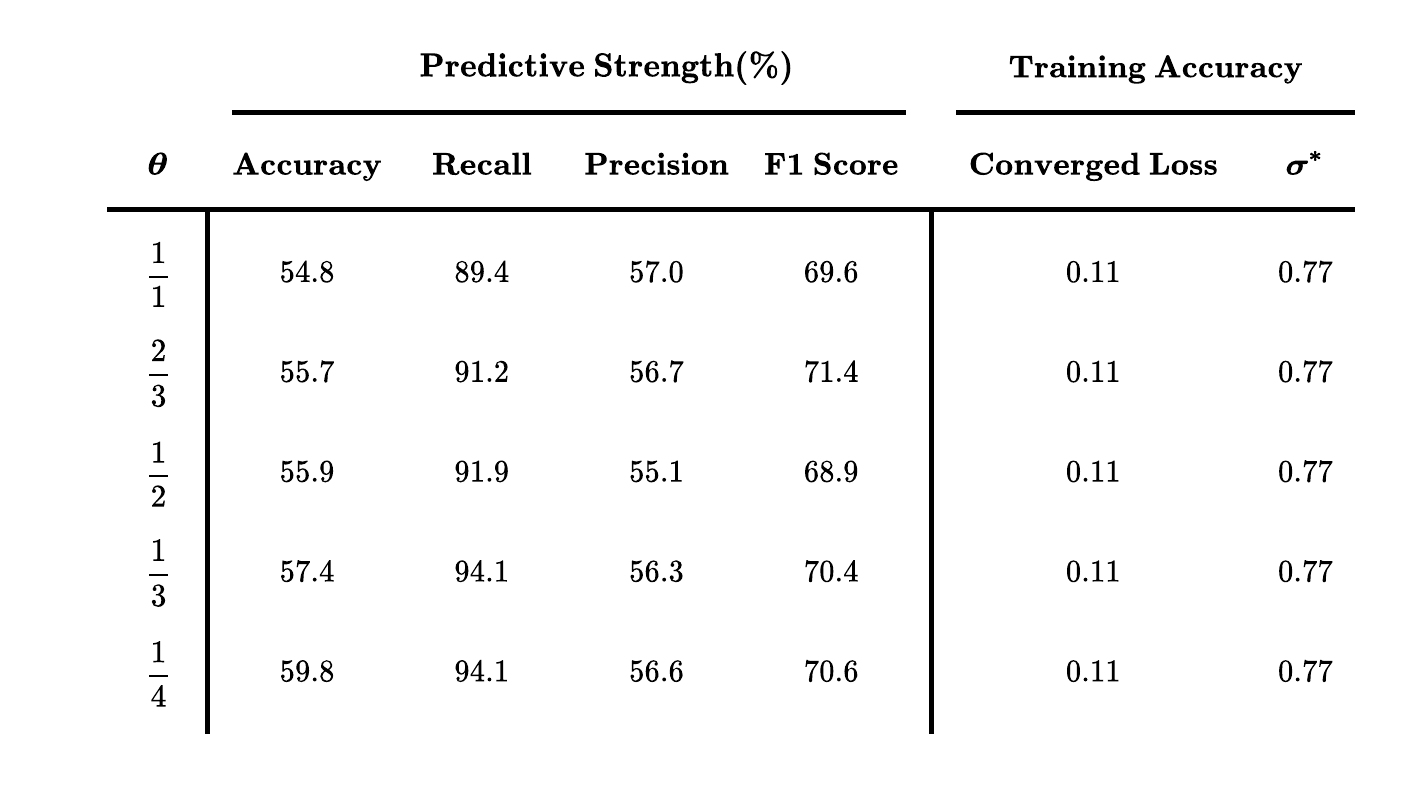}
			
			\label{table2}
		\end{adjustwidth}
		
	\end{table}

	\begin{table}[!htb]

		\begin{adjustwidth}{-1.0in}{2.3in}
			\caption{
				{\bf Model performance - BTC/USDT}}
			\includegraphics[width=\textwidth]{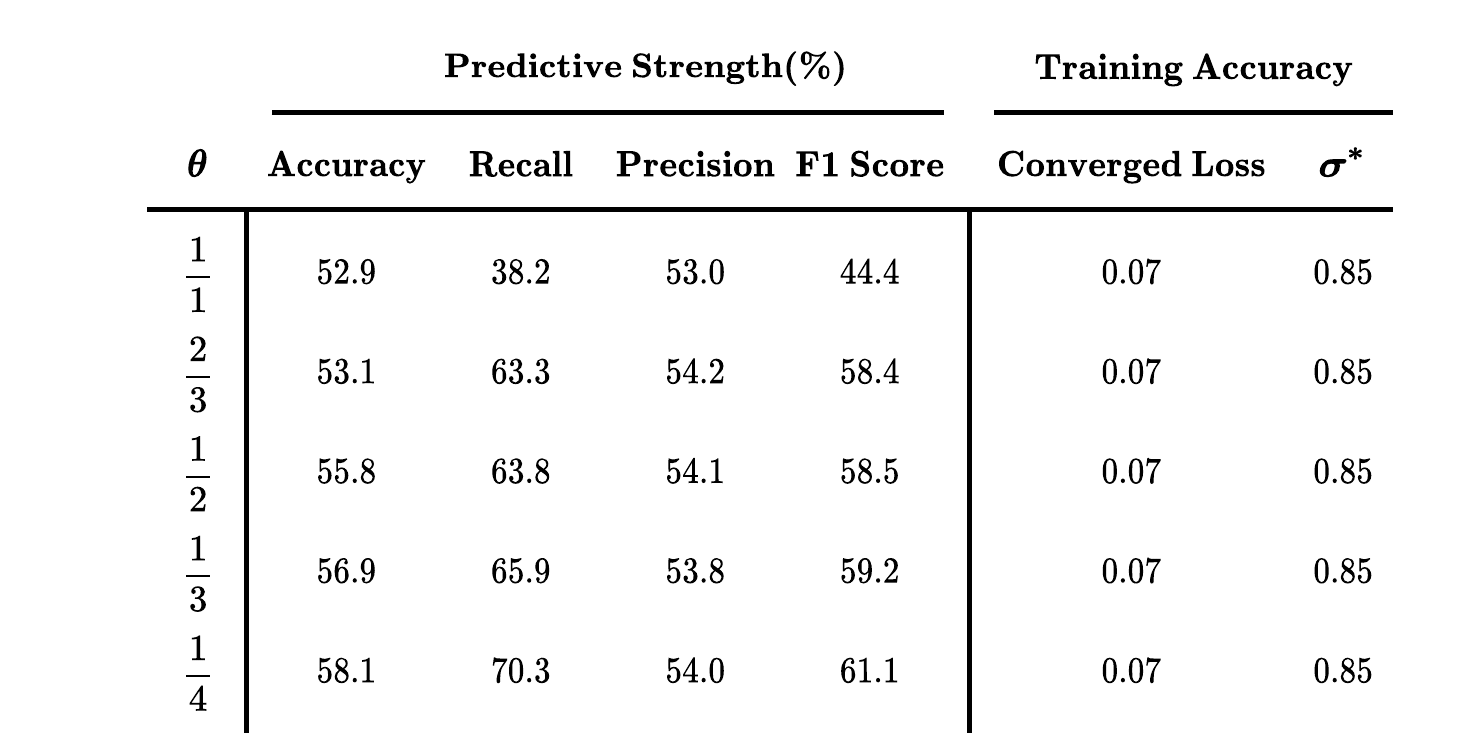}
			\label{table3}
		\end{adjustwidth}
		
	\end{table}

	\begin{table}[!htb]

		\begin{adjustwidth}{-1.0in}{2.3in}
			\caption{
				{\bf Model performance - ETH/USDT}}
			\includegraphics[width=\textwidth]{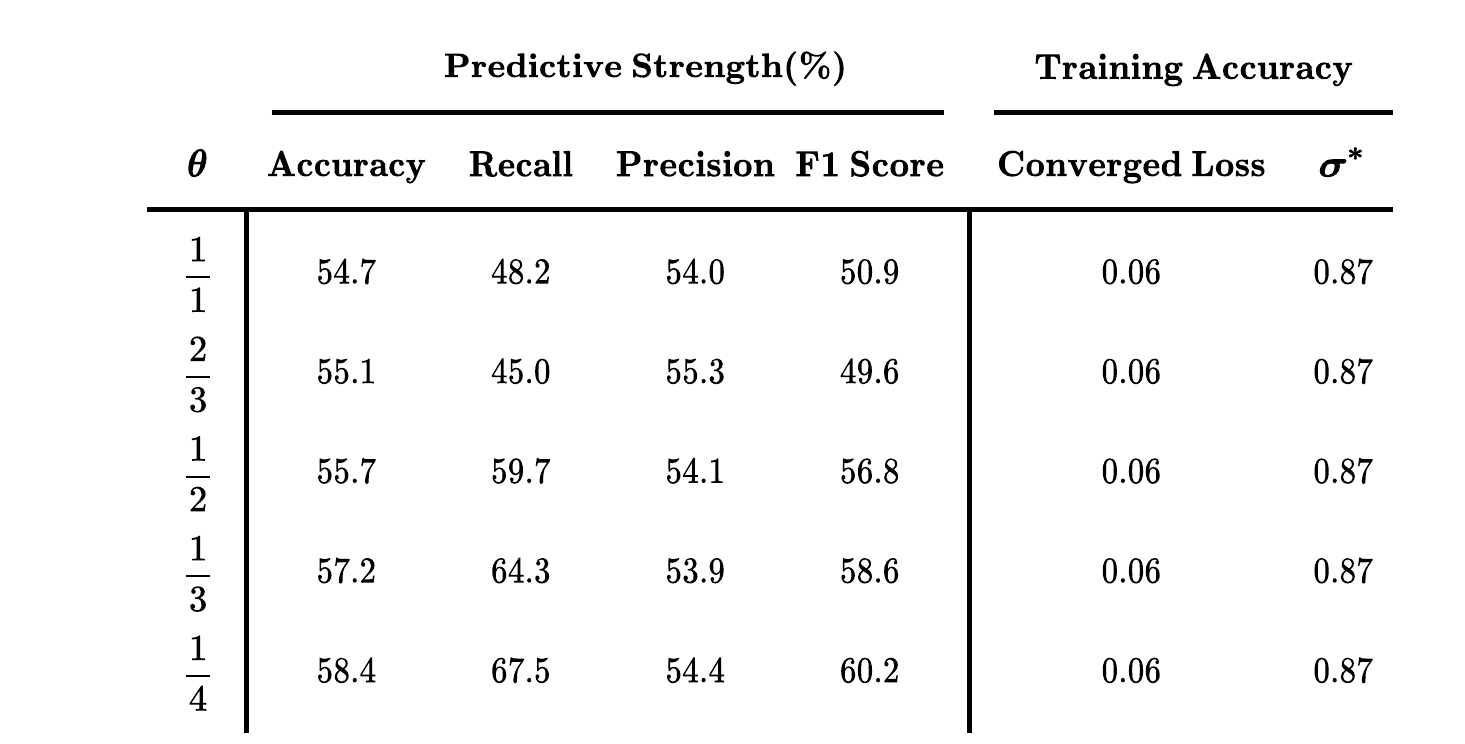}
			
			\label{table4}
		\end{adjustwidth}
	\end{table}

	\subsection*{Trading strategy profitability}
	In order to evaluate the trading strategy profitability, this part reports the total percentage of profit and loss (PnL) in two conditions, without profit saving (consider initial margin plus the amount of profit so far as an accessible margin) and with profit saving (consider just initial margin as an accessible margin). These indicators are provided for different values of $\theta$ according to Table~\ref{table5}, Table~\ref{table6} and Table~\ref{table7}.
	
	\begin{table}[!htb]

		\begin{adjustwidth}{-1.0in}{2.3in}
			\caption{
				{\bf Trading strategy profitability - S\&P500}}
			\includegraphics[width=\textwidth]{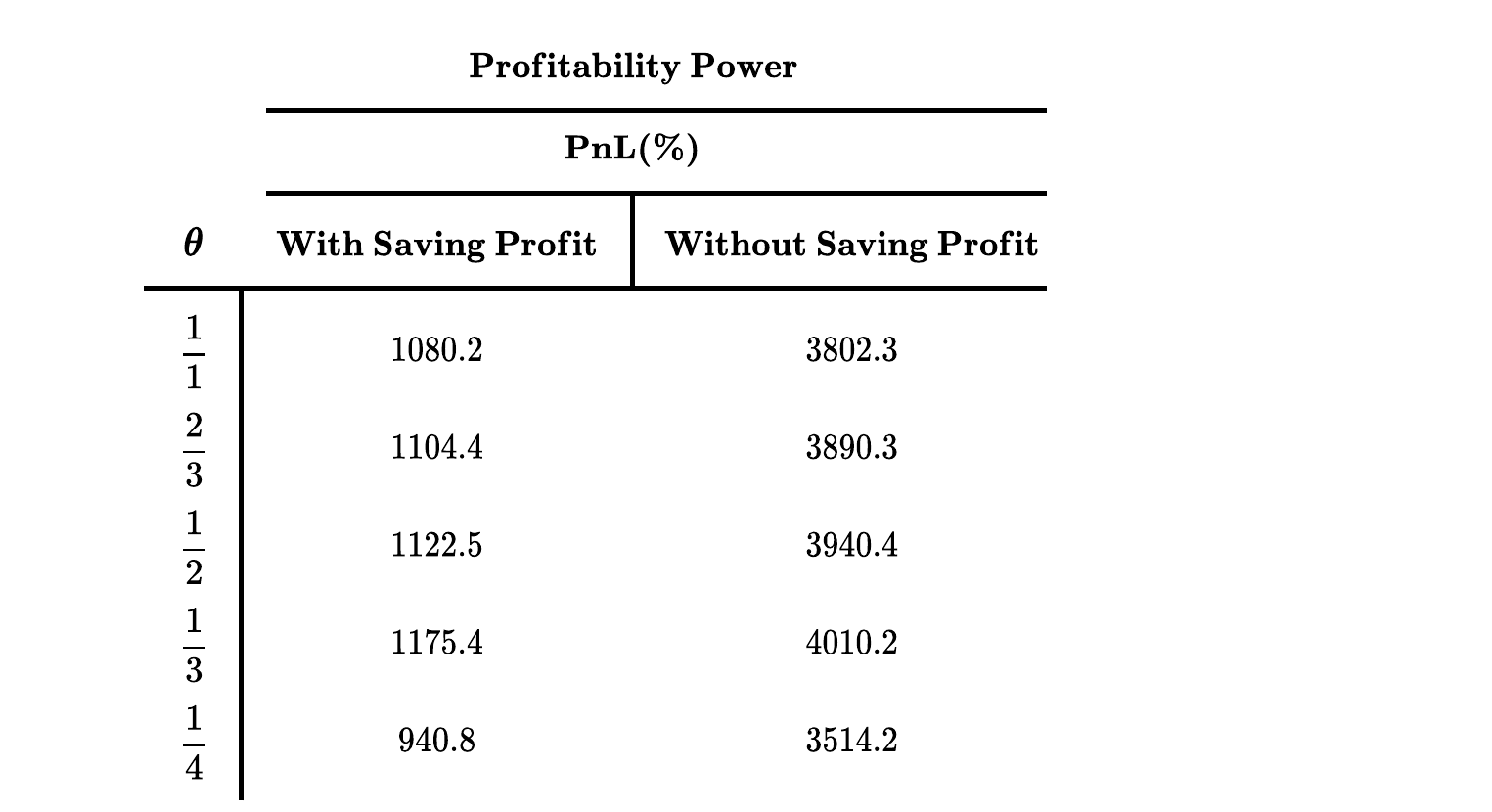}
			
			\begin{flushleft} \small
				 Results show that when we set $\theta$ on 1/4, the profitability of the trading strategy decreases. The reason is that too much sensitivity about confidence deprives us of the opportunity to earn positions with lower amounts of profit.
			\end{flushleft}
			\label{table5}
		\end{adjustwidth}
	\end{table}

	\begin{table}[!htb]

		\begin{adjustwidth}{-1.0in}{2.3in}
			\caption{
				{\bf Trading strategy profitability - BTC/USDT}}
			\includegraphics[width=\textwidth]{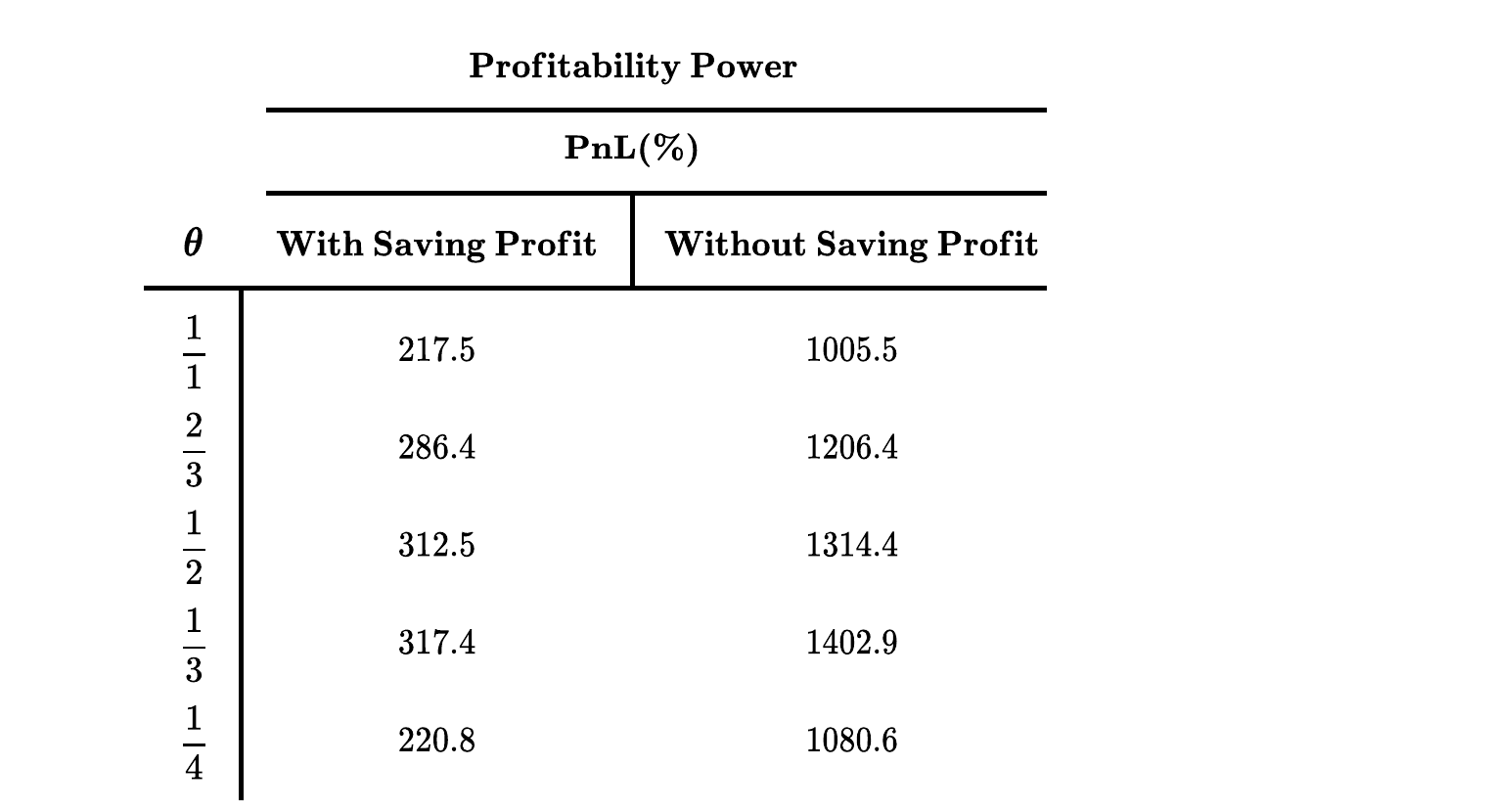}
			
			\label{table6}
		\end{adjustwidth}
	\end{table}

	\begin{table}[!htb]

		\begin{adjustwidth}{-1.0in}{2.3in}
			\caption{
				{\bf Trading strategy profitability - ETH/USDT}}
			\includegraphics[width=\textwidth]{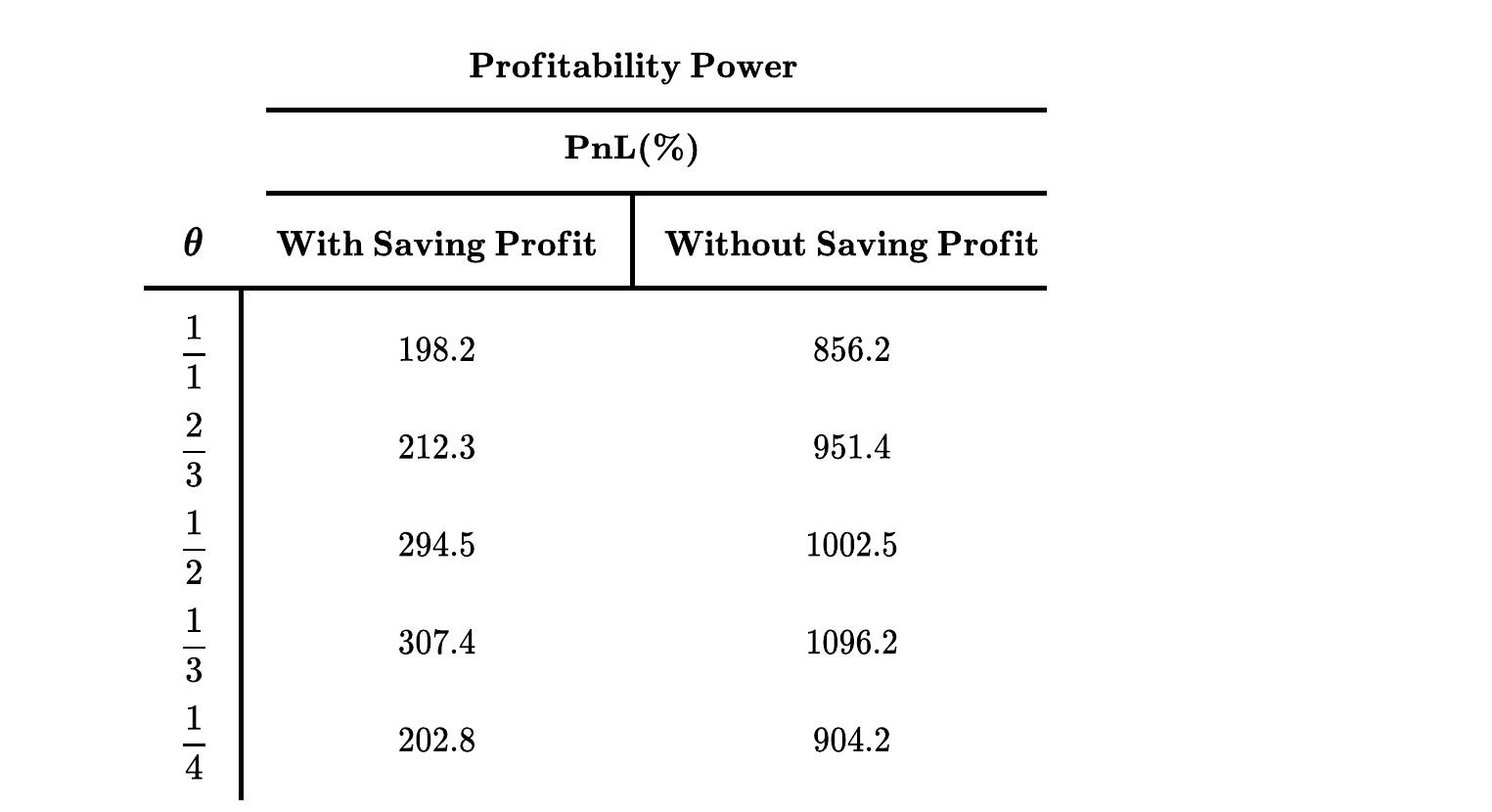}
			
			\label{table7}
		\end{adjustwidth}
	\end{table}

	\section*{Discussion}
	This study takes each step according to an intuition that emphasizes we need to have domain knowledge of the field in which we wish to utilize artificial intelligence. This domain knowledge is exactly what helps us to design each part of our research according to its corresponding reason in the real world.

Introducing a comprehensive framework that cares about all steps from manipulating raw data to defining a trading strategy is a novel research whose main aim is to be implementable in the real world. Using traders' behavior and coming up with corresponding elements in our framework is the main edge of this research. It avoids supposing unrealistic assumptions and tries to suggest solutions to help organizations, researchers, or people who would like to earn profit from their financial activities.

With respect to all complexity in financial markets, too many other topics, components, and problems have remained unexplored, and further research is required to address them. This research is the first attempt to adopt a compatible mindset with financial markets and define a guideline to interpret these time series data better. Further research will also be considered to address and solve other obstacles we may encounter in financial markets.

	\section*{Conclusion}
	This research provides a novel framework to manipulate financial time series data and exploit deep neural networks to forecast the one-step-ahead closing price. It also recommends defining a trading strategy to be able to implement our work in real-world conditions. The first part of this study, the data preparation, consists of six important steps. Firstly it generates some financial features then denoises the data. Since there is no evidence on how important these features are, the feature engineering part is considered as the third step. Having done that, it splits data into three categories and creates a representative dataset. Finally, it uses auto-encoders to extract high-level features and perform dimensionality reduction. The second part of this study, the modeling, aims at 
	designing an intuitive neural network architecture by which data instances are classified. This part consists of three elements. The first element, the Conv element, aims at changing data instances' resolution to have a more robust prediction. The second element, the LSTM element, is designed to transform the information from the past to take the price path into account when acting on the current candle. Finally, the FC element is considered to binary-classify the LSTM outputs. The last part of this study, the trading strategy, talks about how to the trained model. It mentions some unrealistic assumptions and suggests solutions to deal with them. Finally, it reviews indexes and indicators that we should use to evaluate the work.

The performance of this framework is tested on three different pieces of data, including two coins from the cryptocurrency market and S\&P500 market data. The result shows that not only is this model able to provide a robust prediction, but it can also provide a considerable amount of profit such that our balance would increase 10x-12x during one year.
	
	\bibliographystyle{apalike}
	\bibliography{Refs.bib}
	
\end{document}